\ifx\pdftexversion\undefined\else\pdfoutput=1\fi
\documentclass[10pt]{article} % For LaTeX2e
\usepackage[preprint]{tmlr}
\usepackage{amsmath,amsfonts,bm}

\def\eqref#1{equation~\ref{#1}}
\def\1{\bm{1}}

\DeclareMathAlphabet{\mathsfit}{\encodingdefault}{\sfdefault}{m}{sl}
\SetMathAlphabet{\mathsfit}{bold}{\encodingdefault}{\sfdefault}{bx}{n}

\newcommand{\E}{\mathbb{E}}

\newcommand{\R}{\mathbb{R}}

\DeclareMathOperator*{\argmin}{arg\,min}

\usepackage{hyperref}
\usepackage{url}
\usepackage{graphicx}
\usepackage{booktabs}
\usepackage{multirow}
\usepackage{amsmath,amssymb,amsthm}
\usepackage{xcolor}
\usepackage{array}
\usepackage{tabularx}

\newtheorem{proposition}{Proposition}

\newcommand{\nom}{nearest-of-$m$}
\newcommand{\Nom}{Nearest-of-$m$}
\newcommand{\imlewm}{IMLE-WM}
\newcommand{\ms}[2]{$#1 \pm #2$}
\newcommand{\best}[1]{\textbf{#1}}

\newcommand{\deff}{d_{\mathrm{eff}}}

\newcommand{\IS}{\mathrm{ES}}
\graphicspath{{figures/}}

\title{How Many Candidates Does a One-Step World Model Need?\\A Random-Codebook Account of \Nom{} Training on Frozen DINO Tokens}

\author{\name Yong Hwi Song \email thddydgnl1937@gmail.com \\
      \addr School of Information and Communication Engineering\\
      Chungbuk National University, Cheongju, Republic of Korea
      \AND
      \name Keon Myung Lee \email kmlee@cbnu.ac.kr \\
      \addr School of Computer Science\\
      Chungbuk National University, Cheongju, Republic of Korea}

\def\month{MM}  % Insert correct month for camera-ready version
\def\year{YYYY} % Insert correct year for camera-ready version
\def\openreview{\url{https://openreview.net/forum?id=XXXX}} % Insert correct link to OpenReview for camera-ready version

\begin{document}

\maketitle

\begin{abstract}
One-step stochastic world models avoid the mode averaging of deterministic feature predictors without the tens of network evaluations per step that diffusion and flow matching require. The simplest such objective is \nom{} matching (conditional implicit maximum likelihood estimation, IMLE): draw $m$ candidate futures and pull only the one closest to the observed outcome. Recent work conjectures that a small $m$ suffices for feature-space world models and, conversely, that a strictly proper energy score should replace the hard minimum. We test both claims with one-step world models on frozen DINOv2/v3 patch tokens, using benchmarks whose futures can be enumerated. Under mode-like uncertainty, \nom{} with $m{=}16$ covers the enumerated futures better than 50-step flow matching, one-step flow generators and an energy-score generator at matched parameter counts, with about $30\times$ lower latency than 50-step flow matching, and its rollouts stay on the token manifold where flow rollouts drift. The candidate budget is governed by one quantity: the \nom{} objective is the expected distortion of a random codebook of $m$ draws from the generator. This view yields three predictions, which we recorded before the corresponding runs: (i) memorization of the realized outcome in contexts that never repeat, whose strength is governed by the codebook's miss \emph{distortion} $I(m)$ rather than by its miss probability, and which an energy-score generator does not escape by raising $m$; (ii) a flattening of mode probabilities with $m$ that matches a known optimum within $0.01$; and (iii) under-dispersion on continuous high-dimensional outcomes, which bounds the method on real robot video. Per-step matching on self-rollouts cuts long-horizon error by a third on a stochastic Push-T variant, and an ablation separates its manifold benefit from its coverage benefit. Planning success does not improve despite accurate risk estimates, a negative result that we report together with its diagnosis.
\end{abstract}

\section{Introduction}
\label{sec:intro}

World models that predict the future in the feature space of a frozen vision foundation model are cheap to train and plan with \citep{zhou2025dinowm,baldassarre2025dinoworld,assran2025vjepa2}, but a deterministic predictor returns the conditional mean, which lies between the modes of a multimodal future \citep{porcher2026flowwm}. Stochastic feature-space world models fix this with diffusion or flow matching at tens of network evaluations (NFE) per step \citep{porcher2026flowwm,boduljak2025vfmf,walker2025generalist}, a cost that matters when a planner imagines thousands of futures per decision. Several recent works therefore build one-step stochastic world models \citep{lu2026driftworld,agibot2026geact,lee2026beckmann,park2026priming,hsu2026vsf}.

A simpler one-step objective predates these methods: draw $m$ candidate futures from a noise-conditioned generator, find the one closest to the observed outcome, and pull only that one towards it. This objective is conditional implicit maximum likelihood estimation \citep[IMLE;][]{li2018imle,li2019diverse}; it is also the min-of-$K$ objective of explorative modeling \citep[XM;][]{gladstone2026xm}, whose $K$ is the budget we call $m$, and a sampled relative of winner-takes-all learning \citep{guzman2012mcl,rupprecht2017mhp}. Because every data point is explained by at least one sample, mode collapse is discouraged by the objective itself, and inference is a single forward pass. IMLE world models beat Gaussian ones on proprioceptive state \citep{aghabozorgi2026wimle}, and \citet{gladstone2026xm} conjecture that feature-space world models have few modes, so that a small candidate budget should suffice. In the opposite direction, \citet{makabe2026esp} note that a hard minimum over finitely many candidates is not a strictly proper scoring rule and show that an energy-score generator covers a two-branch toy distribution better than min-of-$K$.

This paper asks a quantitative version of both claims: \emph{how many candidates does a one-step \nom{} world model need, when is no budget enough, and is a strictly proper score the better objective?} We answer with a conditional IMLE world model on frozen DINOv2/v3 patch tokens, compared at matched parameter counts against deterministic regression, flow matching with 50 and 2 steps in the FlowWM recipe, two one-step flow generators (improved MeanFlow and Shortcut models) and an energy-score generator, on benchmarks whose futures can be enumerated, so that coverage is measured rather than inferred. One quantity organizes the results: the \nom{} loss is, in expectation, the distortion of quantizing the outcome with a \emph{random codebook} of $m$ draws from the generator. This is a sampled version of the known equivalence between winner-takes-all learning and conditional quantization \citep{rupprecht2017mhp,letzelter2024wta,perera2024amcl}, and in the continuous limit it coincides with the Gaussian rate--distortion test channel \citep{shannon1959coding,cover2006elements}. The mathematics is classical; our contribution is to show that this single correspondence predicts, in advance of the runs, where the budget matters and where it cannot help.

\paragraph{Contributions.}
(1) \textbf{A controlled test of the small-budget conjecture.} Under mode-like uncertainty, \nom{} with $m{=}16$ covers enumerated futures better than 50-step flow matching (FM-50), one-step flow generators and an energy-score generator at matched parameter counts (coverage $0.93$ vs.\ $0.79$ and $0.36$ on non-repeating contexts; F1 error $7.2$ vs.\ $12.3$\,px on Bouncing Shapes) and at $30\times$ lower latency than FM-50; on a stochastic Push-T variant its precision advantage is significant and its recall advantage consistent but not significant (Section~\ref{sec:results1}).
(2) \textbf{Outcome memorization and its incentive.} In contexts that never repeat, a conditional generator can reach zero training loss by memorizing the realized outcome of each training context, keyed on its appearance. We measure this as the share of samples on the realized mode of training contexts, show that it is governed by the codebook's miss \emph{distortion} $I(m)$ rather than by the miss probability $(1-1/K)^m$ (fit RMSE $0.031$ vs.\ $0.068$ over $K \in \{2,4,8,16\}$), confirm seven of nine predictions recorded before the $K{=}2,16$ runs, and explain why the strictly proper energy score memorizes at every $m$ while \nom{} with $m{=}16$ does not (Section~\ref{sec:results2}).
(3) \textbf{Where no budget suffices.} Mode probabilities flatten with $m$, matching the optimum of \citet{bennett2026freedom} within $0.01$, and the spread on continuous high-dimensional outcomes shrinks to $e_m/\sqrt{d}$ of the true spread, the form of the Gaussian rate--distortion test channel; on BAIR the samples spread along only one to three context-specific directions, and $m{=}64$ does not close the gap to multi-step flow matching on distribution metrics (Section~\ref{sec:results3}).
(4) \textbf{Rollouts and planning.} \nom{} rollouts stay as close to the token manifold as deterministic regression, while flow rollouts drift by up to $4\times$ at $h{=}32$ regardless of the number of integration steps; per-step matching on self-generated contexts \citep[in the lineage of][]{venkatraman2015dad,zhang2025catk} cuts long-horizon recall error by a third on stochastic Push-T, and an ablation separates its manifold benefit from its coverage benefit; planning with \imlewm{} is $50\times$ faster than with FM-50 at equal success, but success does not improve even where the model's risk estimates are accurate (Section~\ref{sec:results4}).

Negative results are reported in Appendix~\ref{app:negative}. The benchmarks are synthetic or simulated except for BAIR, the models are small (8M and 21M parameters), and the flow baselines are far from the scale of \citet{porcher2026flowwm}; Section~\ref{sec:discussion} states these limits.

\section{Related Work}
\label{sec:related}
\textbf{Min-of-$m$ objectives.} IMLE matches each data point to its nearest generated sample \citep{li2018imle}; conditional IMLE ignores its noise input when each condition has a single target \citep{li2019diverse,li2020multimodal}, and later variants reject samples that are already close, anneal a per-sample radius, split the noise hierarchically, or add a robust loss \citep{vashist2024rsimle,aghabozorgi2023adaptive,peng2022chimle,vashist2026roms}. IMLE has been used as a world model on state vectors \citep{aghabozorgi2026wimle}, as a trajectory generator for model-predictive control \citep{lee2026imlempc} and as an action head \citep{rana2025imlepolicy,hosseinkhani2026imlevla}. XM \citep{gladstone2026xm} studies the min-of-$K$ objective as a pretraining axis, includes a one-step world model on Maze2D, and finds that a larger $K$ reduces overfitting on video, which it attributes to expressivity. Conditional min-of-$N$ objectives go back to \citet{fan2017pointset} and \citet{bhattacharyya2018bestofmany}, and multi-hypothesis feature forecasting to \citet{vondrick2016anticipating}. Winner-takes-all and multiple-choice learning \citep{guzman2012mcl,lee2016smcl,rupprecht2017mhp,makansi2019ewta,seo2020tmcl} learn $K$ heads; their optima are conditional quantizers \citep{letzelter2024wta,delattre2025cdq,cortes2025timemcl}, their training follows a rate--distortion curve \citep{perera2024amcl}, and their hard assignment distorts mode weights \citep{wu2026wtagmm}. \citet{bennett2026freedom} derives the optimal mode weights of min-of-$K$, which we confirm; \citet{makabe2026esp} and Branch-JEPA \citep{song2026branchjepa} compare hard assignment with energy-score training.

\textbf{World models in feature space.} DINO-WM, DINO-world and V-JEPA~2 predict frozen features deterministically \citep{zhou2025dinowm,baldassarre2025dinoworld,assran2025vjepa2}; FlowWM \citep{porcher2026flowwm} runs flow matching in DINOv3 feature space and introduces the Bouncing Shapes benchmark that we adopt; VFMF and \citet{walker2025generalist} use diffusion on compressed or frozen features. Recent one-step stochastic world models \citep{lu2026driftworld,agibot2026geact,lee2026beckmann,park2026priming,hsu2026vsf} do not measure mode coverage on enumerable futures, and \citet{lee2026beckmann} report no planning gain on Push-T. That generated latents leave the manifold of clean encoder features is a known problem \citep{zheng2025rae,boduljak2025vfmf,karypidis2026latentforesight}; we measure it for each objective.

\textbf{Memorization and rollout training.} Conditional diffusion models memorize more when the condition is more informative \citep{gu2025memorization,kim2026memorization}; data copying is detected as local over-representation of training points \citep{meehan2020nonparametric,somepalli2023diffusion,carlini2023extracting}; conditional GANs that ignore their noise input exhibit a related collapse \citep{yang2019dsgan,mao2019msgan}; and with unlimited capacity any sample-based objective memorizes \citep{li2018imle}. Our measurement labels outcomes by mode and compares training contexts with held-out contexts. Training on self-generated context is the classical remedy for compounding error \citep{ross2011dagger,venkatraman2015dad,bengio2015scheduled,lamb2016professor,hafner2019planet,chen2024diffusionforcing,huang2025selfforcing,gao2026sword}; closest-among-top-$K$ fine-tuning \citep{zhang2025catk} rolls a tokenized policy out on the candidate nearest the ground truth, and our per-step matching is its continuous-feature analogue. \citet{terver2025drives} and \citet{cheung2026fidelity} find that multi-step rollout losses and one-step fidelity are poor predictors of planning success; our planning results agree.

\section{\Nom{} World Models on Frozen DINO Tokens}
\label{sec:method}

\textbf{Setting.} A frozen encoder maps each observation to patch tokens $s_t \in \R^{N \times D}$ with $N = 256$ tokens on a $16\times16$ grid and $D = 384$ channels: DINOv2 ViT-S/14 at $224$\,px \citep{oquab2024dinov2} for the bounce, push, Push-T and BAIR benchmarks, and FlowWM's DINOv3 feature definition \citep{simeoni2025dinov3,porcher2026flowwm} for Bouncing Shapes. Tokens are precomputed and standardized per token. The model takes a context of three token maps $c_t$, an action $a_t$ and a noise vector $z \in \R^{256}$, and predicts the next map as a residual $\hat{s}_{t+1} = s_t + g_\theta(c_t, a_t, z)$; rollouts feed $\hat{s}_{t+1}$ back into the context, at one network evaluation per step.

\textbf{Generator.} $g_\theta$ is a ConvNeXt stack \citep{liu2022convnext} on the token grid. The context maps are concatenated along channels; the action is injected by feature-wise linear modulation (FiLM); the noise is injected by adaptive layer normalization through a four-layer mapping network, as in ROMS-IMLE \citep{vashist2026roms}; and the output head is a per-token MLP of width 1024, the ``wide head'' that FlowWM found necessary in feature space, which we attach to every method so that parameter counts match within $\pm4\%$. We use two sizes: width 256 with 8 blocks ($8.0$--$8.2$M across methods) for the synthetic benchmarks and width 384 with 12 blocks ($20.6$M) for Push-T and BAIR. The noise enters only the last three blocks (four for the larger model), so the trunk is evaluated once per transition and only the tail once per candidate; this late fusion halves training time at equal coverage (Appendix~\ref{app:impl}).

\textbf{Objective.}
\label{sec:objective}
For each transition we draw $m$ noise vectors (the \emph{candidate budget}, or budget for short), generate $m$ candidates without gradient tracking, select the one nearest to the realized outcome in token space,
\begin{equation}
j^\star = \argmin_{j \in [m]} \big\| s_{t+1} - \hat{s}^{(j)} \big\|^2 ,
\label{eq:match}
\end{equation}
and regenerate the selected candidate with gradient tracking, training it with a Geman--McClure loss \citep{geman1987statistical} on its root-mean-square distance $d$: $\rho(d) = 2u^2/(u^2+4)$, where $u = d/(2c)$ and $c$ is the running median of the matching distance. The matching is recomputed for every batch with a per-transition candidate pool \citep[as in WIMLE;][]{aghabozorgi2026wimle}, in bfloat16, without gradient tracking, and in a separate mixed-precision region (Appendix~\ref{app:impl} describes a pitfall of sharing one). The robust loss reproduced the gain reported for ROMS-IMLE in our pilot study but matters little in the final recipe (Appendix~\ref{app:impl}); the budget $m$ is the variable this paper studies. We call the resulting model \imlewm{}. Every 2{,}000 steps, a \emph{memorization monitor} draws 16 samples for each of 200 training contexts and 200 held-out contexts whose next transition is stochastic, and records the share of samples decoded onto the mode that was realized in the data. This share is $1/K$ in expectation on held-out contexts, and it exceeds $1/K$ on training contexts only if the model has learned which mode each training context took.

\textbf{Per-step matching on self-rollouts.}
\label{sec:seqmatch}
Matching whole $H$-step candidate trajectories fails: the number of trajectory modes grows like $K^H$, so the selected candidate is the one that is close to all of them on average (Appendix~\ref{app:negative}). We instead match one step at a time along the model's own rollout. Starting from a real context, at each step $h = 1,\dots,H$ we draw $m$ fresh noise vectors, select the candidate nearest to the true frame $s_{t+h}$ \emph{given the self-predicted context}, train that step, and feed the selected prediction back (detached) as the next context; with $H{=}4$ this costs $mH$ forward passes without gradient tracking and $H$ with it. Section~\ref{sec:seq} compares two alternatives for the fed-back state, a random candidate and the true frame. Selecting the candidate nearest the ground truth is the device of closest-among-top-$K$ rollouts \citep{zhang2025catk}; in \nom{} training this selection is the argmin of the loss itself.

\textbf{Inference, planning and baselines.}
\label{sec:baselines}
One predicted step is one forward pass, and $m_p$ futures for a state are obtained by batching $m_p$ noise vectors. Planning uses cross-entropy-method model-predictive control \citep[CEM-MPC;][]{rubinstein1999cem,zhou2025dinowm}, scoring each candidate action sequence by the mean, over $m_p$ imagined futures, of a distance to the goal; a variance penalty lowered the success rate even with the true simulator as the model (Appendix~\ref{app:negative}). All baselines share the backbone, wide head, context, residual parameterization and training budget unless stated otherwise: \emph{regression} (squared error); \emph{FM-50}, flow matching \citep{lipman2023flow} in the FlowWM recipe (time shift $\alpha = 13$, 50 Euler steps), and \emph{FM-2}, the same model sampled with 2 steps; \emph{iMF} \citep{geng2025meanflow,geng2025imf}, ported from the official code and trained for $3\times$ as many steps; \emph{Shortcut} \citep{frans2025shortcut,hafner2025dreamer4}, trained to the wall-clock budget of \imlewm{} with $m{=}16$; and the \emph{energy score} \citep{gneiting2007strictly} as used by \citet{makabe2026esp}, i.e.\ the \imlewm{} generator trained with the unbiased $m$-sample estimator of $\E\|Y - x\|^\beta - \tfrac{1}{2}\E\|Y - Y'\|^\beta$ ($\beta = 1$, whole-map distance, $m \in \{8, 16\}$), plus a per-token variant (Appendix~\ref{app:impl}).

\section{A Random-Codebook Account of \Nom{} Training}
\label{sec:theory}
Fix a context and write $p$ for the true outcome distribution and $q$ for the generator's distribution. Since the candidates are drawn independently from $q$, the population objective behind Equation~\ref{eq:match} is
\begin{equation}
L(q) = \E_{x \sim p}\, \E_{Y_1,\dots,Y_m \sim q}\, \min_{j \in [m]} \| Y_j - x \|^2 ,
\label{eq:codebook}
\end{equation}
the expected distortion of quantizing $x$ with a random codebook of $m$ i.i.d.\ draws from $q$. Winner-takes-all learning is the deterministic version of this problem, whose minimizers form centroidal Voronoi tessellations of the conditional distribution \citep{rupprecht2017mhp,letzelter2024wta}; for a Gaussian source, the random version is Shannon's random-coding construction at a rate of $\ln m / d$ nats per dimension \citep{shannon1959coding}. The consequences below are elementary (proofs in Appendix~\ref{app:proofs}); their value lies in predictions that we wrote down before the corresponding runs.

\textbf{Outcome memorization in non-repeating contexts.}
\label{sec:theory_memo}
Let the outcome of a context take one of $K$ modes $\mu_1, \dots, \mu_K$. In data whose contexts never repeat, each training context $c_i$ is seen with a single realized mode $\mu_{k_i}$, and two solutions compete for the empirical loss: the \emph{memorizing} solution $q(\cdot\,|\,c_i) = \delta_{\mu_{k_i}}$, which attains zero loss at the cost of storing $k_i$ for every training context (and therefore requires a feature of $c_i$ that identifies it), and the \emph{generalizing} solution $q = p$, which attains
\begin{equation}
I(m) = \E_{k \sim p}\, \E_{S_1,\dots,S_m \sim p}\, \min_{j} \| \mu_{S_j} - \mu_k \|^2 ,
\label{eq:I}
\end{equation}
the distortion of a random codebook of $m$ mode draws.
\begin{proposition}[memorization incentive]
\label{prop:memo}
With non-repeating contexts, the empirical \nom{} loss of the memorizing solution is lower than that of the generalizing solution by exactly $I(m)$. For equiprobable modes with pairwise distances in $[\delta, \Delta]$, $(1 - 1/K)^m \delta^2 \le I(m) \le (1-1/K)^m \Delta^2$, so the incentive vanishes exponentially once $m$ exceeds $K$; if the modes densely fill a curve of length $\ell$ with $K \gg m$, then $I(m) = \Theta(\ell^2/m^2)$, independently of $K$. For squared-error regression the corresponding incentive is $\mathrm{tr}\,\mathrm{Cov}_p$, which no budget reduces.
\end{proposition}
The incentive is a \emph{distortion}, not a probability: a missed mode is weighted by the distance to the nearest drawn mode, and whether $K$ enters at all depends on the geometry. On our bounce benchmark the $K$ modes are equiangular directions on a semicircle, so for $m \lesssim K$ the distortion is nearly the same for every $K$ (Appendix~\ref{app:pred} computes it exactly); the rule of thumb ``$m \approx 2K$'' that the two values $K \in \{4, 8\}$ alone would suggest is a coincidence, and Section~\ref{sec:collapse} rejects it.

\textbf{High-dimensional continuous outcomes.}
\label{sec:theory_shrink}
\begin{proposition}[isotropic shrinkage]
\label{prop:shrink}
Let $p = \mathcal{N}(\mu, \sigma^2 I_d)$ and $q = \mathcal{N}(\mu, s^2 I_d)$. For large $d$, $L(q) \approx d\sigma^2 + d s^2 - 2 s \sigma \sqrt{d}\, e_m$ with $e_m = \E\max_{j \le m} \xi_j$, $\xi_j \sim \mathcal{N}(0,1)$, so the optimal spread is $s^\star/\sigma = e_m/\sqrt{d}$ ($e_4 = 1.03$, $e_{16} = 1.77$, $e_{64} = 2.34$; $e_m \approx \sqrt{2 \ln m}$ for large $m$).
\end{proposition}
This is the first-order form of the reproduction variance $\sigma^2(1 - m^{-2/d})$ of the Gaussian rate--distortion test channel \citep{cover2006elements}, and the correspondence is exact in construction because IMLE draws its codebook i.i.d.\ from the reproduction distribution. It gives a dichotomy between $\ln m$ and the effective dimension $\deff$ of the conditional outcome: a few discrete modes are covered once $\ln m > \ln K$ (the regime of Proposition~\ref{prop:memo}, in which an i.i.d.\ min-of-$N$ loss over-disperses low-dimensional densities; \citealp{thiede2019variety}), whereas when $\deff \gg e_m^2$ the samples contract towards the mean and their spread grows only like $\sqrt{\ln m}$.

\textbf{Flattening of mode probabilities.}
\label{sec:theory_flatten}
Even when every mode is covered, the population optimum for equidistant modes satisfies $p_k (1 - q_k)^{m-1} = \mathrm{const}$ \citep[Theorem~7]{bennett2026freedom}: at $m{=}1$ it collapses onto the most frequent mode, at $K{=}2$ and $m{=}2$ it is calibrated, and for large $m$ the differences $q_k - q_l \approx \ln(p_k/p_l)/(m-1)$ vanish, so rare modes are over-represented; for equiprobable modes the optimum is exactly $q = p$ for every $m \ge 2$ \citep[Theorem~4]{bennett2026freedom}. The budget therefore trades coverage against calibration, which benchmarks with equiprobable modes cannot reveal (Section~\ref{sec:flatten}).

\textbf{The energy score's incentive does not depend on $m$.}
\label{sec:theory_es}
\citet{makabe2026esp} prefer the energy score $\IS_\beta(q, x) = \E_{Y \sim q}\|Y - x\|^\beta - \tfrac{1}{2}\E_{Y, Y' \sim q}\|Y - Y'\|^\beta$, $\beta \in (0,2)$, because it is strictly proper. Strict propriety is a population property; in the finite-sample regime above, the energy score has a further consequence.
\begin{proposition}[memorization incentive of the energy score]
\label{prop:es}
(a) $\IS_\beta(q, x)$ is half the energy distance between $q$ and $\delta_x$ \citep{dawid2007geometry,gneiting2007strictly}, so with non-repeating contexts the unique minimizer of the empirical risk is the memorizing solution, for every $m \ge 2$. (b) The generalizing solution attains $\tfrac{1}{2}\E_{Y,Y' \sim p}\|Y - Y'\|^\beta$, so the incentive is half the expected outcome spread and does not depend on $m$ ($m$ changes only the variance of the estimator). (c) For $\beta = 2$ the score reduces to squared-error regression. (d) A per-token score averaged over the token grid inherits (a) and (b).
\end{proposition}
Normalizing each incentive by the incentive of its fit term alone ($\E\|Y - x\|^\beta$ for the energy score, $I(1)$ for \nom{}) gives a unit-free comparison: regression and the energy score sit at $1/2$ for every $m$, whereas \nom{} sits at $I(m)/I(1)$, which is $0.52$, $0.21$, $0.054$ and $0.005$ at $m = 2, 4, 8, 16$ for $K{=}4$ on the bounce geometry. Parts (a)--(c) are classical \citep{gneiting2007strictly,dawid2007geometry}, and \citet{li2018imle} note the unconditional case of (a); the comparison and its empirical test (Section~\ref{sec:es}) are new. A second difference concerns the \emph{direction} of the spread: with an unknown bias of length $a$ and an isotropic spread $s$ in $d$ dimensions, the energy score is minimized at $s = a$ independently of $d$ (a special case of Proposition~4.1 of \citealp{debortoli2025distributional}), whereas Proposition~\ref{prop:shrink} gives $s \approx a\,e_m/\sqrt{d}$. The energy score thus calibrates the size of the spread but is insensitive to its direction in high dimensions \citep{chakraborty2021framework,pinson2013discrimination,scheuerer2015variogram,marcotte2023regions}, so an isotropic spread leaves the token manifold, whereas \nom{} under-disperses but stays on it (an interpretation; Appendix~\ref{app:proofs}).

The predictions these propositions make, and where each is tested, are listed in Appendix~\ref{app:pred}.

\section{Experimental Setup}
\label{sec:setup}
\begin{table}[t]
\caption{Benchmarks. ``Contexts repeat'' means the same context appears in the training data with several outcomes, which removes the possibility of memorizing the realized outcome. Details in Appendix~\ref{app:bench}.}
\label{tab:bench}
\begin{center}
\footnotesize
\setlength{\tabcolsep}{4pt}
\resizebox{\linewidth}{!}{%
\begin{tabular}{lllllc}
\toprule
Name & Source & Stochasticity & Repeat & Actions & Model \\
\midrule
A1 Bouncing Shapes & FlowWM generator & normal bounce vs.\ reversal per collision & yes & none & 8M (joint 27--29M) \\
A2-div / A2-fix bounce & ours & $K \in \{2,4,8,16\}$ directions at a wall; asymmetric $K{=}2$ & no / partly & none & 8M \\
Push (branch risk) & ours & $\pm20^\circ$ (2 modes), ice $\{-60^\circ,0,60^\circ\}$ (3), absorbing pit & no & 2-D & 8M \\
B stochastic Push-T & DINO-WM data, re-simulated & hidden center of gravity, 4 modes & no & 10-D & 21M \\
BAIR robot pushing & \citet{ebert2017bair} & continuous, high-dimensional & no & 4-D & 21M \\
\bottomrule
\end{tabular}}
\end{center}
\end{table}
\textbf{Benchmarks} (Table~\ref{tab:bench}; Appendix~\ref{app:bench}). \emph{A1}: the public FlowWM generator, used unchanged \citep{porcher2026flowwm}: a square and a circle that, on hitting a wall, bounce normally or reverse their velocity with probability one half; $16{,}384$ training clips (one eighth of FlowWM's) from $5{,}324$ initial conditions with $3.08$ futures each on average, so contexts repeat; FlowWM's DINOv3 features, protocols and heatmap readout. \emph{A2}: a ball moving in a $64\times64$ frame whose direction after hitting a wall is drawn uniformly from $K$ equiangular directions (about $28\%$ of transitions are stochastic; no actions). A2-div redraws the background color, ball color, radius, speed and two distractor rectangles for every episode, so the nearest training context of a test context is $1.9\times$ farther away than the typical nearest neighbor within the training set and every training context carries an identifying key; A2-fix keeps the appearance fixed ($0.35\times$). Each setting has $40$k training frames and 400 held-out wall contexts; an asymmetric $K{=}2$ variant has mode probabilities $(0.8, 0.2)$. \emph{Push}: the same renderer with a 2-D push action deflected by $\pm20^\circ$ in the open region (two modes) and by $\{-60^\circ, 0, 60^\circ\}$ on an ice patch beside an absorbing pit (three modes); in the \emph{branch-risk task} the path to the goal runs along the pit, so one of the three ice modes leads into it. \emph{B}: the Push-T action sequences of DINO-WM \citep{zhou2025dinowm} replayed in the simulator with the T block's center of gravity hidden at one of four positions per episode (mass and friction have no effect under quasi-static pushing); $2{,}400$ training episodes; a multi-future test set of 128 pre-contact starting states $\times$ 64 futures $\times$ 16 steps. \emph{BAIR}: $64\times64$ frames with 4-D actions; $4{,}096$ training and 256 evaluation sequences with 16 samples each.

\textbf{Metrics.} \emph{Coverage} and \emph{precision}: 16 samples per held-out stochastic context are decoded by a position readout (test error $0.13$\,px; Appendix~\ref{app:readout}) and assigned to the nearest enumerated mode within a tolerance; coverage is the fraction of modes hit, and precision is the fraction of samples that land on a mode (an ideal 16-sample sampler reaches coverage $1 - (1-1/K)^{16}$). \emph{Memorization index}: the same share, on the realized mode of 200 training contexts (chance $1/K$; \emph{excess memorization} is $(\text{index} - 1/K)/(1-1/K)$). \emph{kNN precision/recall} \citep{kynkaanniemi2019improved} on PCA-reduced tokens; \emph{deterministic-transition error}: the readout error on non-stochastic transitions. \emph{FlowWM precision/recall error} (A1, B): with twice as many rollouts as true futures per starting state, the mean distance from each rollout to the nearest true future (precision), from each true future to the nearest rollout (recall), and their harmonic mean (F1); L1 over the shape centers on A1, L2 over the eight T vertices on B. \emph{Manifold distance}: the distance from a predicted token map to the nearest real one, divided by the median nearest-neighbor distance among real maps ($1$ = on the data). \emph{Best-of-$n$}: the readout (or token) error of the sample nearest the truth; \emph{spread}: the mean pairwise distance among the samples. On BAIR we also report token RMS errors, the spread-to-error ratio, and the Fr\'echet distance and kNN recall of pooled tokens. \emph{Latency} is measured per predicted step at batch size 1; \emph{planning} reports success rate, pit rate and wall-clock time per MPC step.

\textbf{Protocol.} All models use AdamW (learning rate $2\times10^{-4}$, $\beta_2 = 0.9$), batch size 128, an exponential moving average of the weights, and 12k training steps unless stated otherwise (8k for the budget grid, the asymmetric variant and per-step matching; 36k for iMF; 24k for Shortcut); parameter counts match within $\pm4\%$. Main comparisons use three seeds, grids and follow-up runs two, and ablations one; uncertainty is quantified with bootstrap intervals over contexts and, for Push-T, a hierarchical bootstrap over seeds and starting states (Appendix~\ref{app:stats}). All experiments ran on two RTX A5000 GPUs, about 300 GPU-hours in total. The predictions for the $K{=}2,16$ grid, the energy-score follow-up runs and the fed-back-state ablation were written down before those runs and are reproduced in the appendix, including the ones that missed.

\section{Results I: Does a Small Candidate Budget Suffice?}
\label{sec:results1}

\begin{table}[t]
\caption{A2-div ($K{=}4$, randomized appearance), 8M models, 12k steps, three seeds (energy score $m{=}16$: two). Ideal-sampler coverage $0.99$; memorization chance $0.25$; latency per step at batch 1 on a shared RTX A5000.}
\label{tab:a2}
\begin{center}
\scriptsize
\setlength{\tabcolsep}{3.5pt}
\begin{tabular}{lcccccccc}
\toprule
Method & NFE & Coverage $\uparrow$ & Precision $\uparrow$ & kNN recall $\uparrow$ & Det.\ err.\ (px) $\downarrow$ & Memo.\ index & ms/step & Train (h) \\
\midrule
Regression & 1 & \ms{0.15}{0.01} & \ms{0.61}{0.02} & 0.00 & \ms{1.09}{0.03} & \ms{0.99}{0.01} & 4.4 & 0.3 \\
FM-50 (FlowWM recipe) & 50 & \ms{0.79}{0.01} & \ms{0.51}{0.00} & \ms{0.91}{0.00} & \ms{2.24}{0.02} & \ms{0.29}{0.01} & 107 & 0.4 \\
FM-2 (same model) & 2 & \ms{0.60}{0.00} & \ms{0.46}{0.01} & \ms{0.56}{0.01} & \ms{1.72}{0.01} & \ms{0.29}{0.01} & 6.5 & 0.4 \\
iMF ($3\times$ steps) & 1 & \ms{0.56}{0.01} & \ms{0.23}{0.02} & \ms{0.92}{0.03} & \ms{14.5}{0.7} & \ms{0.25}{0.01} & 3.4 & 3.2 \\
Shortcut ($2\times$ steps) & 1 & \ms{0.66}{0.00} & \ms{0.42}{0.01} & \ms{0.77}{0.01} & \ms{3.60}{0.07} & \ms{0.27}{0.01} & 4.3 & 0.7 \\
Energy score, $m{=}8$ & 1 & \ms{0.36}{0.02} & \ms{0.61}{0.02} & \ms{0.77}{0.06} & \ms{1.06}{0.04} & \ms{0.87}{0.06} & 3.1 & 1.2 \\
Energy score, $m{=}16$ & 1 & \ms{0.28}{0.01} & \ms{0.61}{0.01} & \ms{0.60}{0.02} & \ms{1.06}{0.03} & \ms{0.97}{0.01} & 3.3 & 2.2 \\
\imlewm{}, $m{=}4$ & 1 & \ms{0.46}{0.01} & \ms{0.68}{0.01} & \ms{0.65}{0.01} & \ms{1.14}{0.07} & \ms{0.75}{0.01} & 3.7 & 0.6 \\
\imlewm{}, $m{=}16$ & 1 & \best{\ms{0.93}{0.01}} & \best{\ms{0.82}{0.01}} & \best{\ms{0.94}{0.01}} & \ms{1.55}{0.05} & \ms{0.29}{0.00} & 3.3 & 1.1 \\
\bottomrule
\end{tabular}
\end{center}
\end{table}

\textbf{Non-repeating contexts (A2-div).}
\label{sec:a2}
Table~\ref{tab:a2} is the main comparison (Figure~\ref{fig:scatter}, Appendix~\ref{app:more}). \imlewm{} with $m{=}16$ covers $0.93$ of the enumerated modes, against an ideal-sampler ceiling of $0.99$, with precision $0.82$: its samples sit on the modes. FM-50 reaches a coverage of $0.79$ with precision $0.51$, scattering its samples between the modes; FM-2, iMF and Shortcut are below it on both measures. The energy-score generator has the \emph{lowest} coverage of all stochastic methods ($0.36$ at $m{=}8$, $0.28$ at $m{=}16$) and, like regression, has memorized the training contexts (memorization index $0.87$--$0.97$), while \imlewm{} with $m{=}4$ lies in between (coverage $0.46$, index $0.75$); Section~\ref{sec:results2} analyzes this column. The cost of $m{=}16$ is a larger error on deterministic transitions ($1.55$ vs.\ $1.14$\,px for $m{=}4$) and $1.7\times$ the training time of $m{=}4$; the inference cost is independent of $m$, at $3.3$\,ms per step against $107$\,ms for FM-50.

\begin{table}[t]
\caption{Bouncing Shapes F1 error in px (frame $256$\,px), three seeds. Autoregressive: 3-frame context, 16 one-step predictions, 8M models. Joint: 16 frames predicted at once from 16 context frames, 27--29M models. A last-frame-copy baseline scores $21.7$ in every column; iMF, Shortcut and the energy score reach $16.8$, $12.1$ and $8.9$ (autoregressive, probe; Appendix~\ref{app:readout}). \citet{porcher2026flowwm} report $4.31$ for FlowWM and $17.1$ for their deterministic baseline under their heatmap readout with eight times our training data.}
\label{tab:a1}
\begin{center}
\scriptsize
\begin{tabular}{lccccc}
\toprule
 & & \multicolumn{2}{c}{Autoregressive} & \multicolumn{2}{c}{Joint 16$\to$16} \\
\cmidrule(lr){3-4}\cmidrule(lr){5-6}
Method & NFE & probe & heatmap & probe & heatmap \\
\midrule
Regression & 1 & \ms{8.09}{0.15} & \ms{8.41}{0.15} & \ms{5.14}{0.01} & \ms{22.48}{0.36} \\
FM-50 & 50 & \ms{12.28}{0.13} & \ms{14.11}{0.20} & \ms{3.86}{0.02} & \ms{12.59}{0.14} \\
FM-2 & 2 & \ms{9.95}{0.19} & \ms{12.49}{0.38} & \ms{3.83}{0.02} & \ms{14.80}{0.17} \\
\imlewm{}, $m{=}4$ & 1 & \best{\ms{6.61}{0.18}} & \best{\ms{6.83}{0.16}} & \best{\ms{3.30}{0.02}} & \ms{11.73}{0.31} \\
\imlewm{}, $m{=}16$ & 1 & \ms{7.21}{0.10} & \ms{7.64}{0.18} & \ms{3.32}{0.07} & \best{\ms{9.04}{0.06}} \\
\bottomrule
\end{tabular}
\end{center}
\end{table}

\textbf{Bouncing Shapes (A1).}
\label{sec:a1}
Table~\ref{tab:a1} covers two prediction protocols (autoregressive one-step rollouts, and FlowWM's joint prediction of 16 future frames from 16 context frames) and two readouts (our probe and FlowWM's heatmap readout, which we ported). In all four combinations \imlewm{} has the lowest F1 error, by at least ten seed standard deviations: $6.6$ vs.\ $12.3$\,px for FM-50 in the autoregressive protocol, and $9.0$ vs.\ $12.6$\,px under FlowWM's readout in FlowWM's own protocol. Contexts repeat here, so there is nothing to memorize and $m{=}4$ suffices. The readout matters: the deterministic joint model moves from $5.1$ to $22.5$\,px when the readout changes, which accounts for most of the gap to the $17.1$\,px that \citet{porcher2026flowwm} report for their deterministic baseline, whereas the autoregressive models barely move because they emit a clean feature map at every step (Section~\ref{sec:manifold}). Appendix~\ref{app:more} reports the baseline fairness checks (longer training preserves the ranking, iMF saturates after 24k steps, Shortcut is trained to \imlewm{}'s wall-clock budget, parameter counts match within $\pm4\%$). The caveat is scale: our joint FM-50 reaches $12.6$\,px under FlowWM's readout, where the FlowWM paper reports $4.31$\,px with eight times the data, so our claims are confined to matched small-scale compute.

\begin{table}[t]
\caption{Stochastic Push-T multi-future test (128 pre-contact starts $\times$ 64 futures $\times$ 16 steps): precision / recall error over the eight T vertices in px (frame $512$\,px), 21M, three seeds (iMF two). GT spread: mean pairwise distance among the true futures.}
\label{tab:b}
\begin{center}
\scriptsize
\begin{tabular}{lcccccc}
\toprule
 & & \multicolumn{2}{c}{$h{=}8$} & \multicolumn{2}{c}{$h{=}16$} \\
\cmidrule(lr){3-4}\cmidrule(lr){5-6}
Method & NFE & precision & recall & precision & recall \\
\midrule
GT spread (reference) & & \multicolumn{2}{c}{30.8} & \multicolumn{2}{c}{48.3} \\
Regression & 1 & \ms{24.8}{1.1} & \ms{39.6}{3.4} & \ms{40.2}{3.0} & \ms{66.4}{8.2} \\
FM-50 & 50 & \ms{27.1}{1.6} & \ms{24.3}{1.8} & \ms{40.7}{1.1} & \ms{42.0}{3.6} \\
FM-2 & 2 & \ms{24.2}{1.7} & \ms{23.3}{1.7} & \ms{39.7}{1.4} & \ms{41.5}{3.8} \\
iMF & 1 & \ms{37.2}{2.6} & \ms{33.5}{3.2} & \ms{55.1}{3.7} & \ms{52.4}{3.4} \\
Shortcut & 1 & \ms{28.0}{1.0} & \ms{24.2}{0.4} & \ms{47.3}{1.9} & \ms{45.9}{1.1} \\
\imlewm{}, $m{=}4$ & 1 & \best{\ms{18.7}{0.4}} & \ms{29.2}{3.4} & \best{\ms{33.8}{3.2}} & \ms{49.1}{3.5} \\
\imlewm{}, $m{=}16$ & 1 & \ms{21.8}{1.8} & \best{\ms{22.1}{2.3}} & \ms{34.1}{0.7} & \best{\ms{37.8}{2.3}} \\
\bottomrule
\end{tabular}
\end{center}
\end{table}

\textbf{Stochastic Push-T (B).}
\label{sec:b}
On the Push-T variant with a hidden center of gravity, \imlewm{} with $m{=}16$ has the lowest recall error at both horizons and a precision error $6.6$\,px below that of FM-50 at $h{=}16$ (Table~\ref{tab:b}). Under a hierarchical bootstrap over seeds and starting states, the precision advantage over FM-50, FM-2 and Shortcut is significant (at $h{=}16$: $-6.6$ $[-8.5, -4.7]$, $-5.6$ $[-7.7, -3.2]$ and $-13.2$ $[-16.3, -10.2]$\,px), whereas the recall advantage, although consistent across the 128 starting states, has an interval that includes zero against the flow baselines ($-4.2$ $[-8.8, +1.1]$ vs.\ FM-50; Appendix~\ref{app:more}). All models mispredict the block's displacement at first contact by about $12$ of $16$\,px, an error that three times more training does not shrink and that masks differences between objectives. \imlewm{} with $m{=}4$ shows the collapse analyzed in Section~\ref{sec:results2}: the best precision but the worst recall among the stochastic models, and a rollout spread of $9$\,px against $22$\,px for $m{=}16$.

\section{Results II: Outcome Memorization and the Candidate Budget}
\label{sec:results2}

\textbf{The phenomenon.}
\label{sec:phenomenon}
Regression reaches a memorization index of $0.99$ on A2-div: it learns which mode every training context took. \imlewm{} with $m{=}4$ climbs to $0.75$ while its held-out coverage falls from $0.77$ at 4k steps to $0.46$ at 12k steps (Figure~\ref{fig:curves}, Appendix~\ref{app:more}); on held-out contexts its index stays at chance, so this is memorization of the realized outcome of each training context, keyed on the context's appearance, rather than a failure to generalize. Training the $m{=}4$ model twice as long, or making it $2.7\times$ larger, raises the index further and lowers coverage (Appendix~\ref{app:more}), so the collapse is not under-training. On A2-fix, where the identifying key is absent, the same $m{=}4$ recipe does not memorize (index $0.26$) and covers $0.97$ of the modes (Table~\ref{tab:es}, Appendix~\ref{app:es}), and with $m{=}16$ the index stays at $0.29$ even when training runs $1.5\times$ longer. The budget, not the recipe, determines whether the model memorizes.

\begin{figure}[t]
\centering
\includegraphics[width=\linewidth]{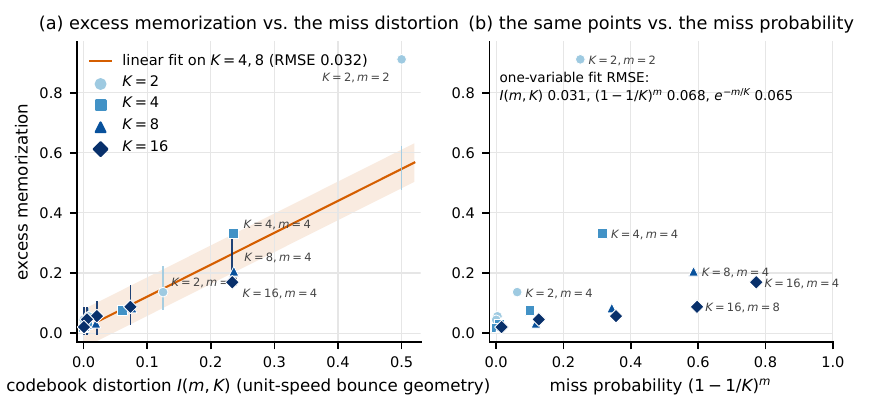}
\caption{Excess memorization on A2-div at 8k steps (two seeds) for $K\in\{2,4,8,16\}$, $m\in\{2,\dots,64\}$. (a) Against the codebook distortion $I(m,K)$ of Equation~\ref{eq:I} (exact for the bounce geometry); line and band: the linear fit on the $K{=}4,8$ points ($\pm2$ RMSE) from which the $K{=}2,16$ predictions were recorded before those runs; vertical bars: the recorded intervals. (b) The same points against the miss probability $(1-1/K)^m$ used by \citet{gladstone2026xm} and \citet{bennett2026freedom}: no collapse.}
\label{fig:collapse}
\end{figure}

\textbf{Excess memorization collapses onto $I(m,K)$.}
\label{sec:collapse}
Proposition~\ref{prop:memo} says that the incentive is the codebook's miss distortion. We trained \imlewm{} for $K \in \{4, 8\}$ and $m \in \{4, 8, 16, 32\}$ (two seeds, 8k steps), fitted excess memorization as a linear function of $I(m,K)$ on those eight points ($0.015 + 1.06\,I$, RMSE $0.032$), recorded predictions with $\pm2$ RMSE intervals for nine runs at $K \in \{2, 16\}$, and only then trained those runs. Seven of the nine fell inside their interval (Figure~\ref{fig:collapse}a; Table~\ref{tab:grid}, Appendix~\ref{app:more}). The two misses lie at the ends of the range and in the directions we had noted as risks: at $K{=}2$, $m{=}2$ ($I = 0.50$, outside the fitted range) memorization is stronger than the linear extrapolation ($0.96$ vs.\ $0.77$), and at $K{=}16$, $m{=}4$ it is weaker ($0.22$ vs.\ $0.31$), consistent with a capacity effect, since storing one of 16 outcomes per context costs $\log_2 16$ bits. The alternative hypothesis, that the budget should scale with the number of modes, is rejected: at a fixed $m{=}8$ the excess memorization is $0.075$, $0.083$ and $0.087$ for $K = 4$, $8$ and $16$, as the nearly equal values of $I(8, K)$ on this geometry predict, whereas at a fixed ratio $m/K = 1$ it ranges from $0.91$ to $0.06$; a one-variable linear fit on all points except the extrapolated one has an RMSE of $0.031$ in $I(m,K)$, against $0.068$ in $(1-1/K)^m$ and $0.065$ in $e^{-m/K}$ (Figure~\ref{fig:collapse}b). The linear relation holds for $I \lesssim 0.25$ and saturates above it. The practical rule is that $m$ must make $I(m)$ small relative to the mode spacing on the outcome manifold; here $m = 16$--$32$ brings the excess memorization below $0.05$ for every $K$.

\begin{figure}[t]
\centering
\includegraphics[width=\linewidth]{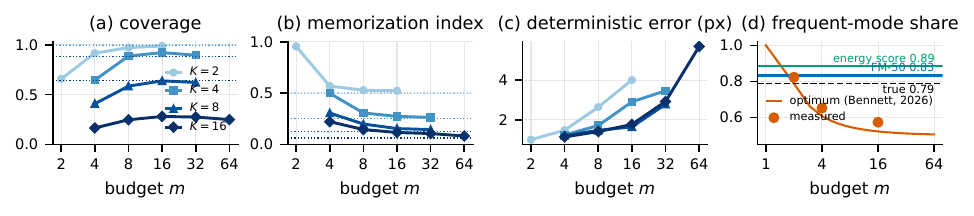}
\caption{What the budget controls (A2-div, 8k steps, two seeds; shade = $K$). (a) Coverage peaks at $m{=}16$ (dotted: ideal sampler) while (b) the memorization index approaches chance (dotted). (c) The deterministic-transition error grows with $m$; training time grows about threefold from $m{=}4$ to $m{=}32$, inference cost not at all (all panels in Appendix~\ref{app:more}). (d) With unequal mode probabilities ($K{=}2$, $p{=}0.79$) the frequent-mode share of \imlewm{} samples falls with $m$ along the optimum of \citet{bennett2026freedom}; FM-50 is calibrated and the energy score over-represents the frequent mode (Section~\ref{sec:flatten}).}
\label{fig:knob}
\end{figure}

\textbf{Countermeasures and the cost of $m$.}
\label{sec:counter}
Raising $m$ is the cleanest way to remove the incentive (Table~\ref{tab:counter}, Appendix~\ref{app:more}): RS-IMLE rejection of candidates already within $\epsilon$ of the target \citep{vashist2024rsimle} changes nothing at $m{=}4$, and Gaussian noise on the context tokens ($\sigma = 1.0$) removes the key and the memorization (index $0.27$) but blurs information the prediction needs (coverage $0.74$, precision $0.61$, deterministic-transition error $1.92$\,px). Figure~\ref{fig:knob} shows the costs of $m$ itself: at 8k steps the deterministic-transition error grows from $1.2$\,px at $m{=}4$ to $2.8$--$3.5$\,px at $m{=}32$ (less with longer training), training time grows about threefold, and coverage and precision saturate at $m{=}16$. The recipe is otherwise insensitive: replacing Geman--McClure by $\ell_2$, or the 256-dimensional noise by a 64-dimensional one, changes coverage and the index by at most $0.01$ (Appendix~\ref{app:impl}).

\textbf{A strictly proper score memorizes regardless of $m$.}
\label{sec:es}
Proposition~\ref{prop:es} predicts that an energy-score generator memorizes in non-repeating contexts at every budget. Trained on A2-div with the \imlewm{} generator unchanged, it does (Table~\ref{tab:a2}; Table~\ref{tab:es} in Appendix~\ref{app:es}): the index reaches $0.87$ at $m{=}8$ and $0.97$ at $m{=}16$ (chance $0.25$); its held-out coverage peaks at $0.50$ near 6k steps and then falls to $0.36$ as the index rises; and a larger $m$ makes it memorize \emph{faster} (index $0.84$--$0.87$ at 8k steps for the two $m{=}16$ seeds, against $0.51$--$0.72$ for the three $m{=}8$ seeds), as expected if $m$ only reduces the gradient variance towards the same minimizer. A per-token energy score, a common choice for continuous tokens, memorizes as well (index $0.97$, coverage $0.22$). The ordering across objectives follows the relative incentives of Section~\ref{sec:theory_es} (regression $0.99$, energy score $0.87$--$0.97$, \nom{} $0.75$ at $m{=}4$ and $0.29$ at $m{=}16$), and with unequal mode probabilities the energy score has no calibration advantage either (frequent-mode share $0.89$ against a true share of $0.79$, coverage $0.58$; Figure~\ref{fig:knob}d).

Memorization requires an incentive, an identifying key and enough revisits of each context. A toy two-branch problem after \citet{makabe2026esp} with a 16-dimensional nuisance ``appearance'' vector in the context reproduces their result when every training context is fresh (the energy score places $0.94$ of its samples on a mode, \nom{} $0.92$; they report $0.962$ and $0.919$), but with $1{,}024$ fixed training contexts the energy score memorizes completely (index $0.999$) and covers both modes on almost no new context, whereas \nom{} stays at $0.51$ and covers $0.99$--$1.00$; removing the appearance vector, or using $16{,}384$ or more contexts, removes the energy score's memorization as well (Appendix~\ref{app:es}). Both conditions transfer to the full environment: the five predictions recorded before the A2-fix and per-token runs (the energy score does not memorize on A2-fix but still trails \imlewm{} in coverage and precision; the per-token variant memorizes on A2-div) all held, and on A2-fix the coverage gap shrinks from $0.57$ to $0.04$ while a precision gap of $0.14$ remains. A second cause is spread off the manifold. On Bouncing Shapes, where nothing can be memorized, the energy score still trails \imlewm{} and regression (F1 $8.9$ vs.\ $7.2$ and $8.1$\,px); its error rises to $33.8$\,px under FlowWM's clean heatmap readout and falls back to $11.7$\,px under a readout trained with feature noise $0.5$, whereas \imlewm{} and regression move by less than $0.5$\,px; and on A2-div its token-space spread exceeds that of \imlewm{} while its readout spread is less than half (Appendix~\ref{app:readout}). The diversity is spent in directions that readouts do not see, as the direction-insensitivity argument suggests. The conclusion is conditional: with fresh contexts or without an identifying feature the energy score is competitive, but in non-repeating contexts with identifying appearance, the property that \citet{makabe2026esp} list as a defect of the hard minimum, namely that it is not strictly proper for finite $m$, is precisely what lets \nom{} escape memorization by raising $m$.

\section{Results III: Where No Candidate Budget Suffices}
\label{sec:results3}

\textbf{Mode-probability flattening.}
\label{sec:flatten}
On a bounce variant with $K{=}2$ modes of probability $0.8$ and $0.2$ (realized frequency $0.787$ on the evaluation contexts), the share of the frequent mode among 64 samples per context falls with $m$ as the stationarity condition of \citet{bennett2026freedom} prescribes: $0.82$, $0.65$ and $0.57$ for $m = 2$, $4$ and $16$, against theoretical values of $0.79$, $0.61$ and $0.52$ (Figure~\ref{fig:knob}d; Table~\ref{tab:flatten}, Appendix~\ref{app:more}). Every method, including FM-50 ($0.83$), measures about $0.04$ above its reference value, a common offset of the readout-based assignment; relative to FM-50 the measured flattening is $-0.18$ and $-0.26$, against theoretical $-0.18$ and $-0.27$. The trade-off is real: $m{=}2$ is calibrated but memorizes (index $0.93$) and covers only $0.75$; $m{=}16$ covers $0.99$ but reports the modes as nearly equiprobable; FM-50 does both; and the energy score does neither (share $0.89$, coverage $0.58$). Inverting the stationarity condition to recover $p$ is ill-conditioned for large $m$ (errors in the log-odds are amplified about $60\times$ at $m{=}16$), so a large budget is a coverage device, not a probability estimate; for equiprobable modes, the calibration is exact (Section~\ref{sec:planning}).

\textbf{Under-dispersion on real robot video (BAIR).}
\label{sec:bair}
BAIR \citep{ebert2017bair} has no enumerable modes; its uncertainty is a continuous, high-dimensional variation of the arm and the objects. \imlewm{} is the most accurate model there (best-of-16 token error $0.527$ at $h{=}16$ against $0.843$ for FM-50; manifold distance $1.30$, like regression, against $1.78$) but loses clearly on distribution metrics (Fr\'echet distance $2.09$ vs.\ $1.38$ and kNN recall $0.15$ vs.\ $0.84$ for FM-50; Table~\ref{tab:bair}, Appendix~\ref{app:more}). The samples are under-dispersed: their spread-to-error ratio at $h{=}1$ is $0.22$ at $m{=}4$ and $0.45$ at $m{=}16$, the same on training and evaluation contexts, so this is not memorization. Raising the budget to $m{=}64$ moves the ratio only to $0.54$ at $2.4\times$ the training time, and matching in a 512-dimensional pooled space lowers it to $0.31$--$0.38$; no one-step \imlewm{} variant closes the gap. Proposition~\ref{prop:shrink} reproduces the growth of the spread with $m$ using one fitted parameter ($\deff \approx 27$, fitted at $m{=}16$, predicts $0.27$ and $0.58$ at $m = 4$ and $64$ against measured values of $0.22$ and $0.54$; Figure~\ref{fig:bair}a, Appendix~\ref{app:more}), but our recorded attempt to remove that parameter failed: the participation ratio of the regression residuals is about $1{,}030$, far outside a factor of two of $27$, and would predict a spread five times smaller than observed. An exploratory analysis designed after this failure suggests why: per context, the samples occupy only $1.2$, $1.8$ and $2.5$ effective directions at $m = 4$, $16$ and $64$; these directions differ across contexts; and they align with the true residual of their own context ten times better than with that of another context, while capturing only $9$--$21\%$ of it (Figure~\ref{fig:bair}b,c). The contraction is therefore a collapse onto a few context-specific directions whose number grows slowly with $m$, the shape of reverse water-filling in rate--distortion theory; with a single true future per context we could not test this quantitatively.

\section{Results IV: Rollouts and Planning}
\label{sec:results4}

\textbf{Rollouts stay on the token manifold.}
\label{sec:manifold}
A one-step model is applied recursively, so beyond one step what matters is whether its outputs stay within the distribution of inputs it was trained on. On A2-div, the manifold distance of \imlewm{} with $m{=}16$ grows from $0.86$ at $h{=}1$ to $1.68$ at $h{=}32$, within $0.1$ of regression ($1.80$), whereas FM-50 reaches $6.16$, FM-2 $6.36$ (so the drift is not an integration error) and iMF $8.04$ (Figure~\ref{fig:manifold}, Appendix~\ref{app:more}); the ordering holds on Bouncing Shapes ($2.08$ vs.\ $2.95$--$3.57$ at $h{=}16$) and on Push-T ($3.28$ vs.\ $4.37$--$5.81$). This explains why the autoregressive models in Table~\ref{tab:a1} are nearly indifferent to the readout while the joint-prediction models lose a factor of $2.7$--$4.4$ under the heatmap readout, and why flow matching has the better best-of-16 error on Push-T alone: its rollouts spread more as they drift ($36$ vs.\ $22$\,px at $h{=}16$), so one of them lands closer (Appendix~\ref{app:more}). A readout trained with feature noise $0.5$ rules out readout fragility as the cause: under it, regression and \imlewm{} improve ($22.5 \to 9.7$ and $9.0 \to 5.3$\,px, joint protocol) while flow matching worsens ($12.6 \to 15.0$).

\begin{table}[t]
\caption{Per-step matching on self-rollouts ($H{=}4$) against one-step matching on stochastic Push-T at an equal training budget (8k steps; 21M, three seeds): precision / recall error (px) and the paired per-seed change.}
\label{tab:seq}
\begin{center}
\scriptsize
\setlength{\tabcolsep}{5pt}
\begin{tabular}{lccc}
\toprule
Stochastic Push-T & $h{=}8$ prec.\ / recall & $h{=}16$ prec.\ / recall & Train (h) \\
\midrule
One-step matching & 25.7 / 25.4 & 41.6 / 43.7 & 1.3 \\
Per-step, feed matched & \best{18.8} / \best{18.4} & \best{24.6} / \best{29.3} & 5.3 \\
Paired change & $-27\%$ / $-28\%$ & $-41\%$ / $-33\%$ & $\times 3.9$ \\
One-step matching, 12k steps & 21.8 / 22.1 & 34.1 / 37.8 & 2.2 \\
\bottomrule
\end{tabular}
\end{center}
\end{table}

\textbf{Per-step matching and what the fed-back state does.}
\label{sec:seq}
Per-step matching with $H{=}4$ reduces the $h{=}16$ recall error on stochastic Push-T from $43.7 \pm 0.9$ to $29.3 \pm 2.8$\,px (hierarchical bootstrap: $-14.3$ $[-19.1, -9.9]$\,px) and the precision error by $41\%$, and it beats the one-step-matched model trained $1.5\times$ longer (Table~\ref{tab:seq}). On the deterministic Push-T control, where the only train--test gap is exposure bias, the best-of-16 error at $h{=}16$ falls by $19\%$, so most of the gain comes from training on self-generated context; on A2-div only the manifold distance improves, and on A1 the best-of-16 rollout error improves by $24\%$ with F1 unchanged (Appendix~\ref{app:seq}). The cost is $3$--$4\times$ the training time. Changing only the fed-back state separates the two mechanisms (Table~\ref{tab:seqabl} and Figure~\ref{fig:seqabl}, Appendix~\ref{app:seq}). Feeding back the true frame, with the same number of matched updates, changes nothing. Feeding back a random candidate keeps the rollout on the manifold at least as well as the matched candidate does ($1.81$ vs.\ $1.92$) but does not improve the best-of-16 error ($22.9$ vs.\ $23.2$) and halves the spread relative to the true futures: a candidate that has followed a different future makes the true next frame an unreachable target, so the one-step distribution widens into jitter that does not accumulate into distinct futures. Only the matched candidate keeps the context on the future that was realized ($1.11\times$ the true spread; best-of-16 error $17.7$\,px). The manifold benefit thus comes from any self-generated context, and the coverage benefit from feeding back the candidate nearest the truth, which in \nom{} training is the argmin of the loss; the ablation was run on Bouncing Shapes only.

\textbf{Planning: equal success, $50\times$ faster, accurate risk estimates.}
\label{sec:planning}
On stochastic Push-T, \imlewm{}, regression and FM-2 are indistinguishable at $32$--$35\%$, FM-50 is at $30\%$ on ten episodes, and the planning time per step is $4.0$\,s for \imlewm{}, half that of FM-2 and $1/50$ that of FM-50 (Table~\ref{tab:plan}, Appendix~\ref{app:plan}); the shared contact error likely prevents every model from exploiting the hidden center of gravity. The branch-risk task isolates the question: the straight path to the goal crosses an ice patch along an absorbing pit, each push on the ice is deflected by $-60^\circ$, $0^\circ$ or $+60^\circ$ with equal probability, and the upward deflection leads into the pit. \imlewm{} estimates the risk correctly (a one-push pit probability of $0.36 \pm 0.03$ against the true $0.33$, and a five-push probability of $0.44$ against $0.43$), FM-2 under-estimates it by a factor of $2.4$ because two Euler steps do not resolve the modes (with 50 steps it gives $0.31$), and regression cannot represent it at all. Yet the \imlewm{} planner succeeds no more often than the regression planner ($55\%$ vs.\ $56\%$) and falls into the pit \emph{more} often ($31\%$ vs.\ $24\%$), while the FM-2 planner, with the worst risk estimates, succeeds most often ($67\%$; the oracle planner reaches $76\%$). Planning performance thus does not follow the model's probabilistic accuracy: a mean-distance cost over a few imagined futures lets the cross-entropy method exploit model errors, and an absorbing failure is not penalized much more than a detour, consistent with \citet{cheung2026fidelity}, \citet{terver2025drives} and \citet{lee2026beckmann}. A planner that uses the estimated mode probabilities directly (through a conditional value-at-risk objective or a failure-probability constraint) is the natural next step; a mean-plus-variance cost is not, since it lowered the success rate from $78\%$ to $50\%$ even with the true simulator as the model (Appendix~\ref{app:negative}).

\section{Discussion and Limitations}
\label{sec:discussion}
\textbf{When to use a one-step \nom{} world model.} If the outcome uncertainty consists of a few discrete alternatives, the random codebook covers them: $I(m)$ vanishes exponentially once $m$ exceeds the number of modes on the outcome manifold, and a budget of $16$--$32$ removes the memorization incentive. If the uncertainty is continuous and high-dimensional, no affordable $m$ covers it, and the samples collapse onto a few context-specific directions. A practical rule is to estimate $I(m)$ from the outcome diversity of near-identical contexts and to pick the smallest $m$ at which it is a few percent of the mode spacing; a hybrid that selects the mode with \nom{} and models the residual with a few-step flow could serve both regimes.

\textbf{Limitations.} (1) The approach under-disperses on real video, and no one-step variant fixed this; the isotropic fit on BAIR uses one free parameter. (2) On stochastic Push-T the advantage over flow matching lies in precision; the recall advantage does not exceed seed variability with three seeds. (3) The benchmarks are synthetic or simulated except for BAIR, the models are small, and the flow baselines are far from the scale of \citet{porcher2026flowwm}. (4) Planning success did not improve on either task. (5) Large budgets flatten mode probabilities, and the energy score did not provide calibration here. (6) The ablations, the BAIR variants and the deterministic Push-T control use a single seed; the $K{=}2,16$ grid and the energy-score follow-up runs use two. (7) Proposition~\ref{prop:es} was derived after we observed the energy score memorizing; only the follow-up runs had predictions recorded in advance, and we examined a single $\beta$ and two distance choices.

\section{Conclusion}
\label{sec:conclusion}
A one-step world model trained with a \nom{} objective on frozen DINO tokens covers enumerable futures better than multi-step flow matching and other one-step generators at matched parameter counts, with one network evaluation per step and rollouts that stay on the data manifold. The budget it needs is set by the distortion of a random codebook of $m$ draws: a few dozen candidates suffice for mode-like uncertainty, and no budget does for continuous high-dimensional uncertainty.

\subsubsection*{Broader Impact Statement}
This work studies the training objectives of small world models on synthetic, simulated and public robot data; we see no risks beyond those of world modeling in general, with one caution: samples from a \nom{} model with a large budget over-represent rare modes (Section~\ref{sec:flatten}) and must not be read as calibrated probabilities in safety-relevant use.

\appendix
\section{Derivations}
\label{app:proofs}

\paragraph{Proposition~\ref{prop:memo}.}
With non-repeating contexts, the empirical loss of a conditional generator decomposes over training contexts. For the memorizing solution every candidate equals the realized mode, so the minimum distance is zero for every context. For the generalizing solution the candidates are $m$ i.i.d.\ draws from $p(\cdot\,|\,c_i)$, and the realized mode is itself a draw from $p$, so the expected minimum distance is $I(m)$ of Equation~\ref{eq:I}. The difference of the two empirical risks is therefore $I(m)$. For the bounds, let the modes be equiprobable with pairwise distances in $[\delta, \Delta]$. The probability that none of the $m$ draws equals the realized mode $k$ is $(1 - 1/K)^m$; on that event the nearest draw is at squared distance in $[\delta^2, \Delta^2]$, and otherwise the distance is zero, which gives the stated bounds. If the modes densely fill a curve of length $\ell$ ($K \gg m$), the draws are $m$ uniform points on the curve and the expected squared arc distance from a uniform point to its nearest draw is $\Theta(\ell^2/m^2)$, as for the nearest of $m$ uniform points on an interval, so $I(m)$ decays polynomially and does not depend on $K$. For squared-error regression the single prediction can memorize the realized mode at zero loss, and the generalizing prediction (the conditional mean) has loss $\mathrm{tr}\,\mathrm{Cov}_p$, independent of any budget. A detail used in Section~\ref{sec:collapse}: the memorization index $s$ is related to the excess memorization by $s = 1/K + (1 - 1/K)\,\mathrm{excess}$, so a $\pm 2$ RMSE interval on the excess scale maps to $\pm 2(1 - 1/K)\,\mathrm{RMSE}$ on the index scale.

\paragraph{Proposition~\ref{prop:shrink}.}
Write $Y_j - X = (\mu - X) + s\,\varepsilon_j$ with $\varepsilon_j \sim \mathcal{N}(0, I_d)$ and $X - \mu \sim \mathcal{N}(0, \sigma^2 I_d)$. Then $\|Y_j - X\|^2 = \|X - \mu\|^2 + s^2 \|\varepsilon_j\|^2 - 2 s\, \varepsilon_j^\top (X - \mu)$. The first term is common to all $j$; the second concentrates at $s^2 d$ with fluctuations of order $s^2 \sqrt{d}$; the third is, conditionally on $X$, Gaussian with variance $4 s^2 \|X - \mu\|^2 \approx 4 s^2 \sigma^2 d$ and independent across $j$, so the minimum over $j$ of the sum is, to leading order in $d$, $d\sigma^2 + d s^2 - 2 s \sigma \sqrt{d}\, e_m$ with $e_m = \E \max_{j \le m} \xi_j$. Minimizing in $s$ gives $s^\star = \sigma e_m / \sqrt{d}$. The spread-to-error ratio used in Section~\ref{sec:bair}, defined as the mean pairwise distance among samples divided by the mean distance from samples to the truth, is $r = \sqrt{2}\rho/\sqrt{1 + \rho^2}$ with $\rho = s/\sigma$ under this model, which is how $\deff$ is fitted from one measured $r$.

\paragraph{Flattening.}
Let the generator place mass $q_k$ on mode $k$ and let all modes be at mutual distance $\Delta$. The probability that none of $m$ draws lands on the realized mode $k$ is $(1 - q_k)^m$, so $L(q) = \Delta^2 \sum_k p_k (1 - q_k)^m$. Minimizing subject to $\sum_k q_k = 1$ gives the stationarity condition $m\, p_k (1 - q_k)^{m-1} = \lambda$ for every $k$ with $q_k > 0$, which is Theorem~7 of \citet{bennett2026freedom}; for $K = 2$ it reads $\big((1-q)/q\big)^{m-1} = (1-p)/p$. If $p_k = 1/K$ for all $k$ the symmetric point $q_k = 1/K$ satisfies the condition and is the unique minimizer of the convex objective for $m \ge 2$ (Theorem~4 of the same work). The numerical optima quoted in the text (for $p = (0.787, 0.213)$: $0.787$, $0.607$, $0.522$ at $m = 2, 4, 16$) solve this equation.

\paragraph{Proposition~\ref{prop:es}.}
(a) The energy distance between $q$ and $\delta_x$ is $2\E_{Y \sim q}\|Y - x\|^\beta - \E_{Y, Y' \sim q}\|Y - Y'\|^\beta - 0$, so $\IS_\beta(q, x)$ is half of it; for $\beta \in (0, 2)$ the energy distance is non-negative and zero only when the two distributions coincide (its square root is a metric on distributions with finite $\beta$-th moments; \citealp{gneiting2007strictly,dawid2007geometry}), so it vanishes only at $q = \delta_x$. The empirical risk over training contexts is a sum of such terms with one $x$ per context, so its unique minimizer is the memorizing solution; the $m$-sample estimator with the unbiased pairwise term has this risk as its expectation for every $m \ge 2$. (b) At $q = p$, $\E_{x \sim p}\,\IS_\beta(p, x) = \E\|Y - x\|^\beta - \frac{1}{2}\E\|Y - Y'\|^\beta = \frac{1}{2}\E\|Y - Y'\|^\beta$ because $x$ and $Y$ are exchangeable. (c) For $\beta = 2$, $\E\|Y - x\|^2 - \frac{1}{2}\E\|Y - Y'\|^2 = \|\E Y - x\|^2$. (d) The per-token score is a sum over tokens of energy scores of the token marginals, and (a)--(b) apply to each term.

\paragraph{Direction insensitivity (interpretation).}
Suppose the generator's mean misses the outcome by an unknown vector of length $a$ and it adds isotropic noise of scale $s$ in $d$ dimensions with $d$ large. Then $\E\|Y - x\|^\beta \approx (a^2 + s^2)^{\beta/2}$ and $\E\|Y - Y'\|^\beta \approx (2 s^2)^{\beta/2}$, and the energy score $(a^2 + s^2)^{\beta/2} - \frac{1}{2}(2s^2)^{\beta/2}$ is minimized at $s = a$ for every $\beta \in (0, 2)$, where it equals $2^{\beta/2 - 1} a^\beta$; this is the $\lambda = 1$ case of Proposition~4.1 of \citet{debortoli2025distributional}, and the gain does not depend on $d$. The same calculation for the \nom{} loss follows Proposition~\ref{prop:shrink} and gives $s \approx a\, e_m/\sqrt{d}$. The approximation requires the bias and the noise to be spread over many dimensions (participation ratio $\gg 1$); the regression residual on BAIR has participation ratio about $1{,}030$, and we did not measure it on the bounce tokens. The argument says nothing about where the isotropic spread points; the claim in Section~\ref{sec:es} that it points off the manifold is an empirical observation.

\section{$I(m,K)$ for the bounce geometry and the recorded predictions}
\label{app:pred}
\label{app:imk}
At a wall frame the $K$ possible next directions are the unit vectors at angles $\theta_k = \big((k + \tfrac12)/K\big)\,180^\circ - 90^\circ$ about the wall normal, $k = 0, \dots, K-1$, with unit speed. For a target mode $k$ let $d^{(k)}_{(1)} \le \dots \le d^{(k)}_{(K)}$ be the sorted squared distances from $\mu_k$ to all modes ($d^{(k)}_{(1)} = 0$). The nearest of $m$ i.i.d.\ uniform draws is the $r$-th closest mode with probability $\big((K - r + 1)/K\big)^m - \big((K - r)/K\big)^m$, so
\begin{equation}
I(m, K) = \frac{1}{K} \sum_{k=1}^{K} \sum_{r=1}^{K} d^{(k)}_{(r)} \Big[ \big(\tfrac{K - r + 1}{K}\big)^m - \big(\tfrac{K - r}{K}\big)^m \Big].
\end{equation}
\paragraph{Predictions tested in the paper.} P1: excess memorization collapses across $K$ against $I(m,K)$, not against $(1-1/K)^m$ ($K{=}2,16$ predictions recorded in advance; Section~\ref{sec:collapse}). P2: the energy score memorizes in non-repeating contexts at every $m$, faster at larger $m$, and stops when contexts carry no identifying feature (Section~\ref{sec:es}). P3: with $p = (0.8, 0.2)$ the frequent-mode share of \imlewm{} samples falls with $m$ as prescribed (Section~\ref{sec:flatten}). P4: on real video the spread grows with $m$ only like $e_m$ and $m{=}64$ does not reach FM-50's distribution metrics (Section~\ref{sec:bair}). P5: the energy score's spread is off-manifold, so its deficit survives a noise-robust readout (Section~\ref{sec:es}).

\paragraph{Recorded predictions for the $K{=}2,16$ grid.} Table~\ref{tab:predfull} lists the values together with the predictions recorded before the $K \in \{2, 16\}$ runs. The predictions were made by fitting excess memorization linearly in $I$ on the eight $K \in \{4, 8\}$ points (excess $= 0.015 + 1.06\,I$, RMSE $0.032$) and mapping to the index scale; the intervals are $\pm 2$ RMSE. We noted two risks at the time: the $K{=}2$, $m{=}2$ point lies outside the fitted range of $I$, and large $K$ might memorize less than predicted because more bits must be stored per context. Both misses are of these kinds.

\begin{table}[h]
\caption{Recorded predictions for the memorization index at 8k steps and the observed two-seed means. The two misses are marked.}
\label{tab:predfull}
\begin{center}
\small
\resizebox{\linewidth}{!}{%
\begin{tabular}{cccccc}
\toprule
$K$ & $m$ & $I(m,K)$ & Chance & Predicted [interval] & Observed \\
\midrule
2 & 2 & 0.500 & 0.50 & 0.77 [0.74, 0.81] & 0.956 (miss) \\
2 & 4 & 0.125 & 0.50 & 0.57 [0.54, 0.61] & 0.568 \\
2 & 8 & 0.008 & 0.50 & 0.51 [0.50, 0.54] & 0.528 \\
2 & 16 & $\approx 0$ & 0.50 & 0.51 [0.50, 0.54] & 0.522 \\
16 & 4 & 0.234 & 0.0625 & 0.31 [0.25, 0.37] & 0.221 (miss) \\
16 & 8 & 0.074 & 0.0625 & 0.15 [0.09, 0.21] & 0.144 \\
16 & 16 & 0.021 & 0.0625 & 0.10 [0.06, 0.16] & 0.115 \\
16 & 32 & 0.005 & 0.0625 & 0.08 [0.06, 0.14] & 0.105 \\
16 & 64 & 0.0006 & 0.0625 & 0.08 [0.06, 0.14] & 0.081 \\
\midrule
4 & 4 / 8 / 16 / 32 & 0.235 / 0.061 / 0.006 / 0.0001 & 0.25 & (fitted) & 0.498 / 0.306 / 0.272 / 0.263 \\
8 & 4 / 8 / 16 / 32 & 0.236 / 0.076 / 0.019 / 0.002 & 0.125 & (fitted) & 0.305 / 0.198 / 0.154 / 0.144 \\
\bottomrule
\end{tabular}}
\end{center}
\end{table}

\section{Benchmark construction}
\label{app:bench}

\paragraph{A2 appearance randomization.}
Each episode of A2-div draws a 22-dimensional appearance vector: background RGB in $[10, 110]$, ball RGB in $[150, 255]$, radius in $[3, 6]$\,px, speed in $[10, 18]$\,px per frame, and two rectangles with top-left corner coordinates in $[0, 48]$, side lengths in $[6, 20]$ and RGB in $[40, 200]$. Frames are rendered at $64\times64$ with an anti-aliased disc, upsampled to $224$\,px and encoded by DINOv2 ViT-S/14. Context novelty is the ratio between the median distance from a test context to its nearest training context and the median nearest-neighbor distance within the training set, both measured in a PCA-64 token space: $1.9$ for A2-div and $0.35$ for A2-fix. The evaluation uses 400 held-out wall contexts; a stochastic transition is one whose current frame touches a wall.

\paragraph{Push environment.}
The $64\times64$ arena contains an ice rectangle ($x \in [20, 44]$, $y \in [14, 40]$) and an absorbing pit ($x \in [24, 40]$, $y \in [0, 14]$). A push action is a 2-D vector; with probability $0.1$ it is zero, and otherwise its direction is uniform and its magnitude is uniform in $[0.3, 1]$. The ball moves $10$\,px along the action direction rotated by the sampled deflection, and it respawns after four frames in the pit. The branch-risk task draws starting positions with $x \in [8, 12]$, $y \in [17, 21]$ and goals with $x \in [52, 56]$ in the same $y$ band. The planner uses CEM-MPC with horizon $T{=}5$, 64 candidate sequences, $m_p{=}8$ imagined futures per candidate (one for regression), and a cost equal to the mean distance between the decoded final position and the goal; 100 episodes per model and seed with a shared set of task seeds. The risk diagnostic evaluates 200 start states with the straight five-push action sequence against 256 true-simulator rollouts per start, and a one-push test from 200 ice states just below the pit.

\paragraph{Stochastic Push-T hidden variable.}
Table~\ref{tab:cog} shows the per-episode randomizations we tried before choosing the center of gravity; mass and friction have no effect under Push-T's quasi-static physics, and damping produces unrealistic sliding. The multi-future test set takes 128 starting states just before the first contact (an earlier version with starts further from contact did not diverge within 8 steps and was discarded), and for each start the 64 futures share the three context frames. Modes closer than $2$\,px in outcome are merged, leaving $3.9$ distinguishable modes per start on average. The CEM planner for Push-T uses $T{=}5$, 64 candidates, $m_p{=}4$, 3 iterations with 8 elites, re-planning every step for at most 10 steps, with the mean token-space $\ell_2$ distance to the goal as the cost, following \citet{terver2025drives}; goals are within 25 steps of the start, and success requires the block within $20$\,px and $20^\circ$ of the goal.

\begin{table}[h]
\caption{Candidate hidden variables for stochastic Push-T: deviation of the T vertices after 16 steps under the same actions (median / maximum over episodes; the nominal displacement is $124$\,px).}
\label{tab:cog}
\begin{center}
\small
\begin{tabular}{lcl}
\toprule
Randomization & Deviation (px) & Decision \\
\midrule
Block mass $0.5$--$2\times$ & 0.0 / 0.0 & no effect (quasi-static) \\
Contact friction $0.3$--$1.2$ & 0.0 / 0.0 & no effect \\
Damping $0$--$0.1$ & 25 / 346 & unrealistic sliding \\
Center of gravity, radius $20$\,px, 4 positions + 3\,px jitter & 28 / 138 & adopted \\
Center of gravity, radius $25$\,px & 56 / 195 & too large \\
\bottomrule
\end{tabular}
\end{center}
\end{table}

\paragraph{BAIR.}
We use the first part of the official archive: $4{,}096$ training sequences and $1{,}024$ held-out sequences (the official test set plus three held-out training files), $64\times64$ frames, 4-dimensional actions, 3 context frames and 16-step rollouts.

\section{Implementation details and ablations}
\label{app:impl}
Code, environments, the queue files that define every run and the per-run outputs behind every figure and table will be released with the paper.

\paragraph{Training.}
AdamW with learning rate $2\times10^{-4}$, $(\beta_1, \beta_2) = (0.9, 0.9)$, no weight decay, a linear warm-up over 1{,}000 steps followed by a cosine decay, batch 128, gradient clipping at norm 1, an exponential moving average of the weights with decay $\min(0.999, (1 + t)/(10 + t))$ used for all evaluations, 12k steps (8k for the budget grid, the asymmetric variant and the per-step-matching comparisons; 36k for iMF; 24k for Shortcut; 48k for the four-times-longer A1 runs). The noise vector has 256 dimensions and is normalized to unit RMS before the mapping network. Late fusion sets the first five of eight blocks (eight of twelve for the 21M model) to a learned constant style. Candidate generation for matching runs without gradient tracking, in bfloat16, in a separate autocast region; if it shares a region with the gradient pass, autocast reuses the weight casts made without gradient tracking and silently cuts the gradient to the weights. The Geman--McClure scale is twice the running median (EMA $0.99$) of the matching distance. The energy-score runs use $\beta = 1$, the whole-map RMS distance, the unbiased pairwise estimator, and backpropagate in two chunks of contexts; the per-token variant computes pairwise differences explicitly to be exact near zero. Flow matching uses the time shift $\alpha = 13$ at training and sampling and the Euler sampler. The iMF port follows the official JAX implementation through a PyTorch re-implementation: dual $u$/$v$ heads of width 640 (to match parameters), the JVP direction taken from the model's own $v$ prediction, and the loss $\|u + (t - r)\,\mathrm{sg}(\partial_t u) - v\|^2$ plus a flow-matching loss on $v$; it converges only with the time-embedding frequency scale set to 1 rather than the usual 1000. Our own earlier MeanFlow implementation diverged in three attempts once the average-velocity term was switched on, and was abandoned. Shortcut models devote a quarter of each batch to the self-consistency target computed with the EMA model. Parameter counts: 8M setting, $7.97$--$8.24$M across methods ($8.24$M for \imlewm{}); 21M setting, $20.6$M for \imlewm{}. Training times in the tables are wall-clock hours on one RTX A5000; some runs shared a GPU and are therefore slower than the stand-alone timings (one training step at batch 128: 178\,ms for $m{=}4$ and 488\,ms for $m{=}16$ with the 8M model; 404 and 1112\,ms with the 21M model).

\paragraph{Recipe choices.}
Table~\ref{tab:recipe} gives the pilot comparisons behind the recipe. The wide head does not change \imlewm{}'s results (it is there to match the flow baselines); late fusion halves training time at equal coverage and lowers the deterministic-transition error; the FlowWM recipe raised the flow baseline substantially. Multi-stage supervision, which ROMS-IMLE found important for images, does not help on a fixed $16\times16$ token grid (Appendix~\ref{app:negative}).

\begin{table}[h]
\caption{Recipe comparisons on A2-div ($K{=}4$, pilot and preparation runs, one seed; bootstrap 95\% intervals over 400 contexts where given). Training hours are on a shared GPU.}
\label{tab:recipe}
\begin{center}
\small
\resizebox{\linewidth}{!}{%
\begin{tabular}{lcccccc}
\toprule
Variant & Coverage & Precision & kNN recall & Det.\ err.\ (px) & Memo.\ (train / val) & Train (h) \\
\midrule
\imlewm{} $m{=}16$, full noise injection, wide head & 0.934 [0.917, 0.949] & 0.83 & 0.94 & 2.31 & 0.29 / 0.24 & 1.93 \\
\imlewm{} $m{=}16$, late fusion, wide head (adopted) & 0.925 [0.907, 0.942] & 0.83 & 0.95 & 1.56 & 0.29 / 0.27 & 1.05 \\
\imlewm{} $m{=}16$, full injection, narrow head (6.6M) & 0.930 [0.909, 0.948] & 0.84 & 0.94 & 2.02 & 0.30 / 0.25 & 1.71 \\
\imlewm{} $m{=}16$, pilot recipe (three spatial stages) & 0.91 [0.89, 0.93] & 0.81 & 0.94 & 2.67 & 0.30 / 0.22 & 3.2 \\
FM, default recipe, 20 Euler steps & 0.58 [0.55, 0.61] & 0.38 & 0.68 & 4.64 & -- & 0.6 \\
FM, FlowWM recipe, 50 steps (adopted) & 0.79 [0.77, 0.82] & 0.51 & 0.91 & 2.27 & 0.29 / 0.24 & 0.4 \\
iMF, time-embedding scale 1000 & \multicolumn{5}{c}{does not converge} \\
iMF, time-embedding scale 1, 12k steps & 0.55 [0.52, 0.58] & 0.22 & -- & 13.1 & -- & 1.0 \\
\bottomrule
\end{tabular}}
\end{center}
\end{table}

\paragraph{Ablations.}
\label{app:ablation}
On A2-div with $m{=}16$ (one seed): replacing Geman--McClure by $\ell_2$ gives coverage $0.93$, precision $0.83$, index $0.28$, deterministic-transition error $1.81$\,px (vs.\ $1.55$); a 64-dimensional noise vector gives $0.92$, $0.83$, $0.28$, $1.50$; training $1.5\times$ longer (18k steps) gives $0.92$, $0.86$, $0.31$, $1.45$. On stochastic Push-T an 8M \imlewm{} with $m{=}16$ reaches recall errors $24.5$ / $39.0$\,px at $h = 8$ / $16$ against $22.1$ / $37.8$ for the 21M model (three seeds), so model size matters little there. In the pilot with the earlier recipe, Geman--McClure improved precision ($0.92$ vs.\ $0.86$) and token-space coverage ($0.85$ vs.\ $0.66$) over $\ell_2$. The iMF learning curve on A2-div (one seed, checkpoints of a 72k-step run) saturates after 24k steps: coverage $0.56$, $0.57$, $0.58$, $0.58$ and precision $0.24$, $0.25$, $0.26$, $0.26$ at 24k, 42k, 60k and 72k steps.

\section{Readouts}
\label{app:readout}
Position readouts are small networks trained on real tokens: a soft-argmax probe for the ball (test error $0.13$\,px on A2-div, $0.05$\,px with DINOv3 on A1), and for Push-T a probe for agent position, block position and angle (T-vertex error $2.0$\,px). Because a probe might decode generated (out-of-distribution) tokens differently from real ones, the pilot cross-checked it against a nearest-neighbor judge that decodes a token map by looking up the nearest of $12{,}000$ real frames in a PCA-128 space: on the fixed-appearance environment the two judges agreed within $0.01$ for every \imlewm{} run (e.g.\ coverage $0.960$ vs.\ $0.960$ at $K{=}4$) and the nearest-neighbor judge was $0.03$--$0.09$ lower for flow matching ($0.841$ vs.\ $0.814$ with 20 steps), so the probe is, if anything, generous to the baselines; the ranking was identical under both. The FlowWM heatmap readout, ported from the public code, maps the $384$-channel token map through a $1\times1$ convolution to 256 channels, a GELU, dropout $0.1$ and a $16\times$ transposed convolution to $256\times256$ logits per shape; it is trained with a pixel-wise cross-entropy and feature noise of standard deviation $0.05$ for 20k steps and reads out the argmax; its error on real features is $0.10$\,px (95th percentile $0.75$\,px). The noise-robust variant is trained with feature noise $0.5$ (error floor $1.03$\,px). Table~\ref{tab:readout} gives all readout combinations for A1, including the energy-score generator, whose rollout manifold distance at $h{=}16$ is $3.05$ against $2.95$--$3.03$ for flow matching and $1.98$--$2.08$ for \imlewm{}.

\begin{table}[h]
\caption{Bouncing Shapes F1 error (px) under three readouts, three seeds (energy score two). The last row is the 48k-step run of Section~\ref{sec:a1} (one seed).}
\label{tab:readout}
\begin{center}
\small
\begin{tabular}{lcccccc}
\toprule
 & \multicolumn{3}{c}{Joint 16$\to$16} & \multicolumn{3}{c}{Autoregressive} \\
\cmidrule(lr){2-4}\cmidrule(lr){5-7}
Method & probe & heatmap & heatmap, noise 0.5 & probe & heatmap & heatmap, noise 0.5 \\
\midrule
Regression & 5.14 & 22.48 & 9.67 & 8.09 & 8.41 & 8.53 \\
FM-50 & 3.86 & 12.59 & 15.00 & 12.28 & 14.11 & 18.38 \\
FM-2 & 3.83 & 14.80 & 16.43 & 9.95 & 12.49 & 16.85 \\
Energy score, $m{=}8$ & -- & -- & -- & 8.94 & 33.77 & 11.65 \\
\imlewm{}, $m{=}4$ & 3.30 & 11.73 & 5.92 & 6.61 & 6.83 & 7.04 \\
\imlewm{}, $m{=}16$ & 3.32 & 9.04 & 5.28 & 7.21 & 7.64 & 7.92 \\
\imlewm{}, $m{=}4$, 48k steps & -- & -- & -- & 4.08 & 4.20 & -- \\
\bottomrule
\end{tabular}
\end{center}
\end{table}

\section{Additional results}
\label{app:more}

\begin{figure}[h]
\centering
\includegraphics[width=\linewidth]{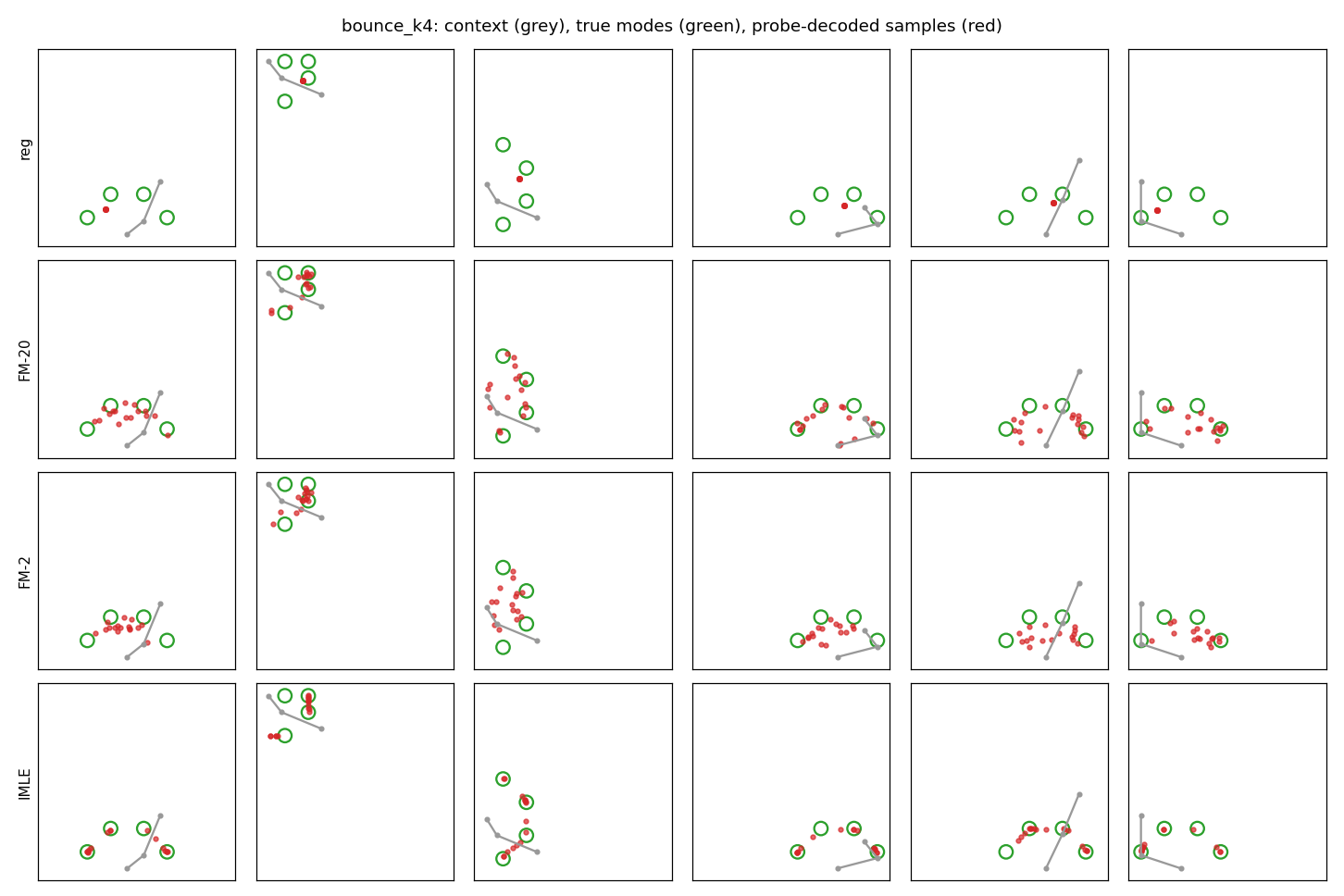}
\caption{Qualitative pilot on the bounce environment with $K{=}4$ modes and fixed appearance (pilot recipe, $m{=}4$). Each column is one context whose ball has just hit a wall: gray, the three context frames; green circles, the four possible next positions; red, 16 samples decoded by the position readout. Regression (top) returns one point near the mean, flow matching with 20 or 2 Euler steps scatters between the modes, and \imlewm{} (pilot recipe) places its samples on the modes.}
\label{fig:scatter}
\end{figure}

\begin{figure}[h]
\centering
\includegraphics[width=\linewidth]{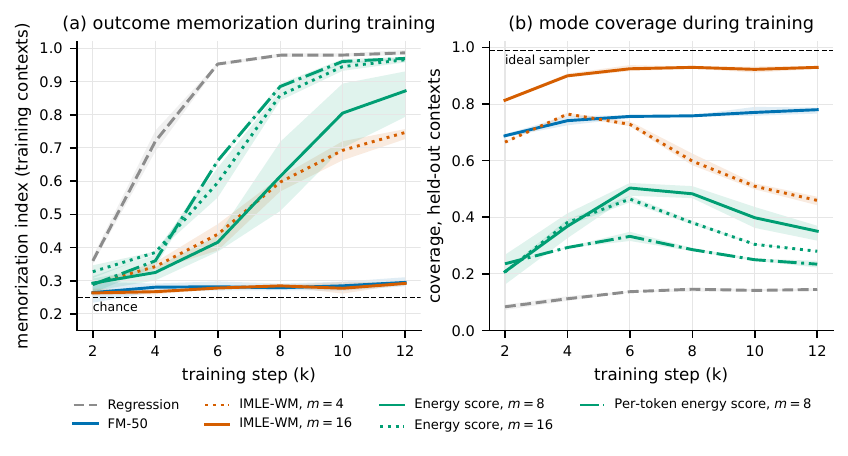}
\caption{Memorization during training on A2-div ($K{=}4$; mean over seeds, band = min--max). (a) Memorization index (share of samples on the realized mode of training contexts); chance is $0.25$ and held-out contexts stay at chance for every method. (b) Coverage on held-out contexts. Regression memorizes completely; \imlewm{} with $m{=}4$ and both energy-score variants memorize progressively and lose coverage as they do; \imlewm{} with $m{=}16$ and FM-50 do not memorize at this budget.}
\label{fig:curves}
\end{figure}

\begin{table}[h]
\caption{Budget grid on A2-div at 8k steps, two seeds: memorization index (chance $1/K$), coverage, precision and deterministic-transition error (px). Coverage is capped by the ideal 16-sample sampler ($1.00$, $0.99$, $0.88$, $0.64$ for $K = 2, 4, 8, 16$). $K{=}16$ is under-trained at 8k steps: its kNN recall rises from $0.12$ to $0.85$ with $m$ while its coverage stays near $0.28$.}
\label{tab:grid}
\begin{center}
\small
\setlength{\tabcolsep}{4pt}
\begin{tabular}{lccccccccccccc}
\toprule
 & \multicolumn{4}{c}{$K{=}2$} & \multicolumn{4}{c}{$K{=}4$} & \multicolumn{4}{c}{$K{=}8$} \\
\cmidrule(lr){2-5}\cmidrule(lr){6-9}\cmidrule(lr){10-13}
$m$ & 2 & 4 & 8 & 16 & 4 & 8 & 16 & 32 & 4 & 8 & 16 & 32 \\
\midrule
Memo.\ index & 0.96 & 0.57 & 0.53 & 0.52 & 0.50 & 0.31 & 0.27 & 0.26 & 0.31 & 0.20 & 0.15 & 0.14 \\
Coverage & 0.66 & 0.92 & 0.97 & 0.99 & 0.65 & 0.89 & 0.92 & 0.90 & 0.41 & 0.58 & 0.64 & 0.62 \\
Precision & 0.78 & 0.85 & 0.85 & 0.68 & 0.69 & 0.76 & 0.77 & 0.72 & 0.54 & 0.62 & 0.64 & 0.60 \\
Det.\ err.\ (px) & 0.99 & 1.47 & 2.64 & 4.01 & 1.23 & 1.70 & 2.91 & 3.46 & 1.20 & 1.46 & 1.64 & 2.78 \\
\midrule
 & \multicolumn{5}{c}{$K{=}16$} \\
\cmidrule(lr){2-6}
$m$ & 4 & 8 & 16 & 32 & 64 \\
\cmidrule(lr){1-6}
Memo.\ index & 0.22 & 0.14 & 0.12 & 0.11 & 0.08 \\
Coverage & 0.17 & 0.25 & 0.28 & 0.27 & 0.25 \\
Precision & 0.32 & 0.38 & 0.41 & 0.38 & 0.33 \\
Det.\ err.\ (px) & 1.13 & 1.40 & 1.76 & 2.92 & 5.71 \\
\bottomrule
\end{tabular}
\end{center}
\end{table}

\begin{figure}[h]
\centering
\includegraphics[width=\linewidth]{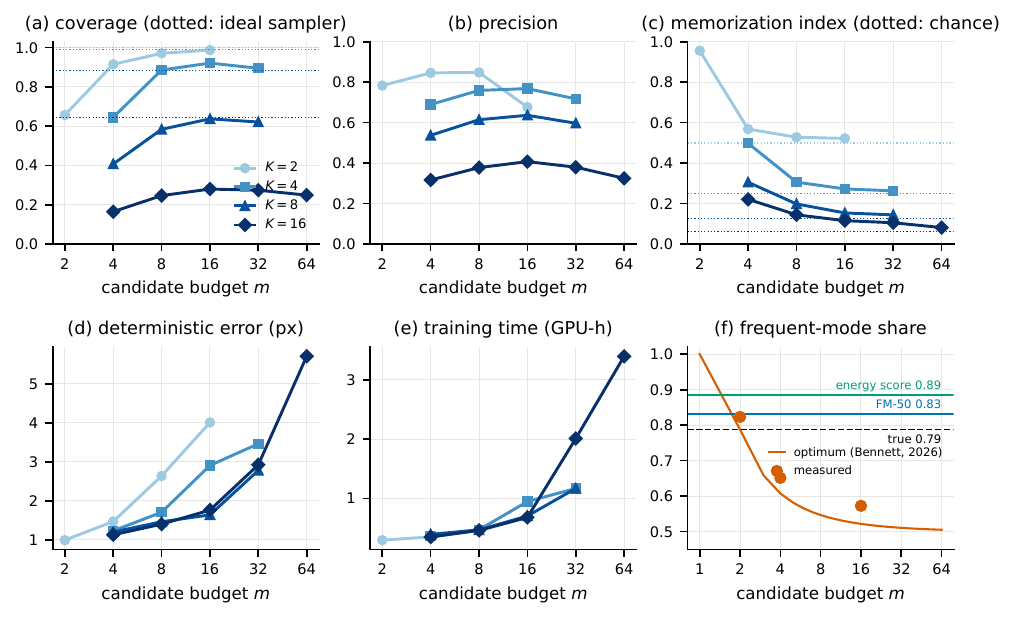}
\caption{Full version of Figure~\ref{fig:knob}: (a) coverage, (b) precision, (c) memorization index, (d) deterministic-transition error, (e) training time (GPU-hours for 8k steps) and (f) frequent-mode share on the asymmetric variant, all against the budget $m$ (A2-div, 8k steps, two seeds).}
\label{fig:knobfull}
\end{figure}

\begin{table}[h]
\caption{Countermeasures against outcome memorization on A2-div ($K{=}4$, 12k steps, two seeds; the $m{=}4$ and $m{=}16$ rows are the three-seed values of Table~\ref{tab:a2}). Memorization index for training / held-out contexts.}
\label{tab:counter}
\begin{center}
\small
\resizebox{\linewidth}{!}{%
\begin{tabular}{lcccccc}
\toprule
Variant & Coverage & Precision & kNN recall & Memo.\ (train / held-out) & Det.\ err.\ (px) & Train (h) \\
\midrule
$m{=}4$ & 0.46 & 0.68 & 0.65 & 0.75 / 0.25 & 1.14 & 0.5 \\
$m{=}4$ + RS-IMLE rejection ($\epsilon{=}0.5$) & 0.46 & 0.69 & 0.65 & 0.74 / 0.25 & 1.17 & 0.5 \\
$m{=}4$ + context noise $\sigma{=}1.0$ & 0.74 & 0.61 & 0.86 & 0.27 / 0.24 & 1.92 & 0.5 \\
$m{=}8$ + context noise $\sigma{=}0.3$ & 0.88 & 0.77 & 0.92 & 0.32 / 0.22 & 1.65 & 0.7 \\
$m{=}16$ & \best{0.93} & \best{0.82} & \best{0.94} & 0.29 / 0.25 & 1.55 & 1.1 \\
\bottomrule
\end{tabular}}
\end{center}
\end{table}

\begin{table}[h]
\caption{Asymmetric bounce ($K{=}2$, $p = 0.787$ on the evaluation contexts, 8k steps, two seeds): share of the frequent mode among 64 samples per held-out context, its theoretical optimum, coverage of both modes, and the memorization index on training contexts (chance $0.5$).}
\label{tab:flatten}
\begin{center}
\small
\resizebox{\linewidth}{!}{%
\begin{tabular}{lccccc}
\toprule
Method & Frequent-mode share & Theory & Share $-$ FM-50 (measured / theory) & Coverage & Memo.\ index \\
\midrule
\imlewm{}, $m{=}2$ & 0.823 & 0.787 & $-0.01$ / $0.00$ & 0.75 & 0.93 \\
\imlewm{}, $m{=}4$ & 0.651 & 0.607 & $-0.18$ / $-0.18$ & 0.96 & 0.68 \\
\imlewm{}, $m{=}16$ & 0.573 & 0.522 & $-0.26$ / $-0.27$ & 0.99 & 0.56 \\
FM-50 & 0.832 & 0.787 & -- & 0.95 & 0.73 \\
FM-2 & 0.915 & -- & $+0.08$ / -- & 0.80 & 0.78 \\
Energy score, $m{=}8$ & 0.886 & 0.787 & $+0.05$ / $0.00$ & 0.58 & 0.91 \\
\bottomrule
\end{tabular}}
\end{center}
\end{table}

\begin{figure}[h]
\centering
\includegraphics[width=\linewidth]{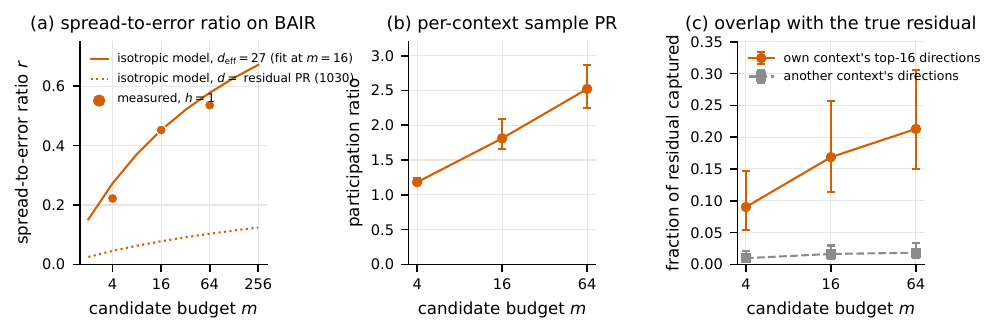}
\caption{Under-dispersion on BAIR ($h{=}1$, 256 evaluation contexts). (a) Spread-to-error ratio against $m$: the isotropic model of Proposition~\ref{prop:shrink} with one parameter fitted at $m{=}16$ ($\deff \approx 27$) reproduces the $m{=}4$ and $64$ points within $0.05$; with $d$ set to the measured participation ratio of the regression residuals ($\approx 1{,}030$) it predicts a spread five times too small. (b) Participation ratio of the 64 samples of a context (median, quartiles): one to three directions. (c) Fraction of the true residual in a context's own top-16 sample directions versus another context's: the directions are context-specific and aligned with the truth, but capture $9$--$21\%$ of it.}
\label{fig:bair}
\end{figure}

\begin{table}[h]
\caption{BAIR robot pushing, 21M models, two seeds (the $m{=}64$ and coarse-matching rows are one seed). Token RMS error of the best of 16 samples and of their mean, manifold distance and spread at $h{=}16$, Fr\'echet distance and kNN recall of mean-pooled PCA-32 tokens against 1{,}024 real frames.}
\label{tab:bair}
\begin{center}
\small
\setlength{\tabcolsep}{3pt}
\resizebox{\linewidth}{!}{%
\begin{tabular}{lccccccccc}
\toprule
Method & NFE & \multicolumn{3}{c}{best-of-16 $\downarrow$ ($h{=}1/8/16$)} & mean $h{=}16$ $\downarrow$ & manifold $h{=}16$ $\downarrow$ & spread $h{=}16$ & FD $h{=}8/16$ $\downarrow$ & recall $h{=}8/16$ $\uparrow$ \\
\midrule
Regression & 1 & 0.268 & 0.481 & 0.556 & 0.556 & \best{1.27} & 0 & 1.44 / 2.65 & 0.21 / 0.08 \\
FM-50 & 50 & 0.338 & 0.681 & 0.843 & 0.885 & 1.78 & 0.94 & \best{0.53 / 1.38} & \best{0.73 / 0.84} \\
FM-2 & 2 & 0.302 & 0.590 & 0.734 & 0.756 & 1.45 & 0.72 & 1.42 / 2.75 & 0.26 / 0.08 \\
Shortcut & 1 & 0.401 & 0.902 & 1.173 & 1.197 & 1.71 & 1.48 & 0.95 / 1.77 & 0.42 / 0.26 \\
\imlewm{}, $m{=}4$ & 1 & 0.260 & 0.460 & 0.532 & \best{0.559} & 1.28 & 0.18 & 1.35 / 2.35 & 0.24 / 0.12 \\
\imlewm{}, $m{=}16$ & 1 & \best{0.258} & \best{0.456} & \best{0.527} & 0.576 & 1.30 & 0.29 & 1.22 / 2.09 & 0.28 / 0.15 \\
\imlewm{}, $m{=}64$ & 1 & 0.257 & 0.461 & 0.534 & 0.589 & 1.33 & 0.32 & 1.23 / 2.00 & 0.28 / 0.14 \\
\imlewm{}, $m{=}16$, coarse matching & 1 & 0.259 & 0.463 & 0.535 & 0.565 & 1.29 & 0.21 & 1.21 / 2.03 & 0.33 / 0.15 \\
\bottomrule
\end{tabular}}
\end{center}
\end{table}

\paragraph{Memorization with longer training and larger models (pilot recipe, A2-div, $K{=}4$, one seed).} With $m{=}4$, doubling the training steps (12k $\to$ 24k) raised the memorization index from $0.66$ to $0.92$ and lowered held-out coverage from $0.52$ to $0.38$; a $2.7\times$ larger model (19M) at 12k steps reached $0.86$ with coverage $0.43$. On held-out contexts the index stayed at chance ($0.23$--$0.25$) in all three runs. With $m{=}16$, training $1.5\times$ longer left the index at $0.31$ (Appendix~\ref{app:impl}).

\paragraph{Baseline fairness checks.}
iMF: a 72k-step run on A2-div (one seed, checkpoints every 6k steps) saturates after 24k steps at coverage $0.56$--$0.58$ and precision $0.24$--$0.26$, so its 36k-step budget in the main tables is not the limiting factor. Flow matching: with four times the steps on A1 (48k, one seed) the F1 error becomes $4.08$ (\imlewm{} $m{=}4$), $7.10$ (FM-2) and $8.99$ (FM-50) under the probe, and $4.20$ for \imlewm{} under the heatmap readout; the FlowWM recipe (wide head, time shift 13, 50 steps) raised the A2-div coverage of our initial 20-step flow baseline from $0.58$ to $0.79$ and its precision from $0.38$ to $0.51$. Shortcut: trained for 24k steps so that its wall-clock time ($0.7$\,h on A2, $1.5$\,h on B) is below that of \imlewm{} with $m{=}16$; sampling with two steps instead of one changed its metrics by less than $0.01$. Parameter counts: $7.97$--$8.24$M (8M setting), $20.6$M (21M setting).

\subsection{Rollout metrics}
\begin{figure}[h]
\centering
\includegraphics[width=\linewidth]{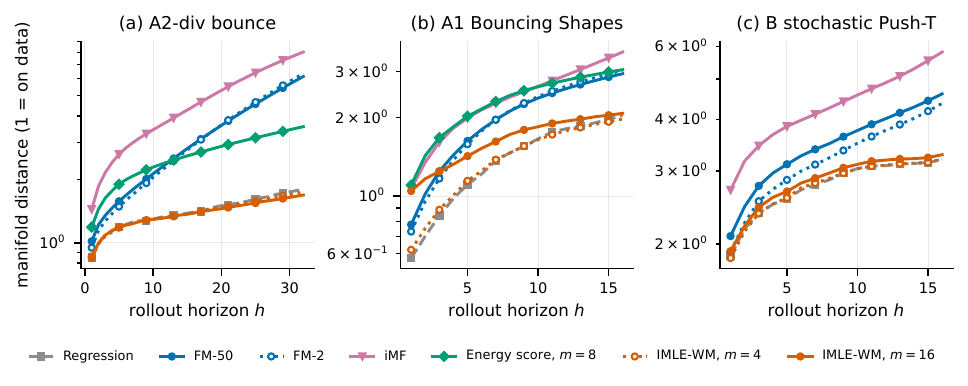}
\caption{Manifold distance of autoregressive rollouts against the horizon (16 rollouts from real contexts, three seeds, log scale; $1$ = on the data). \imlewm{} rollouts stay as close to the data as regression on all three benchmarks; flow-matching rollouts drift equally with 50 and with 2 integration steps; iMF and the energy score drift further.}
\label{fig:manifold}
\end{figure}
Table~\ref{tab:e3} completes Figure~\ref{fig:manifold} with the best-of-16 readout error and the spread at the last horizon.

\begin{table}[h]
\caption{Rollouts from real contexts (16 samples, three seeds; iMF on Push-T two): manifold distance, best-of-16 readout error (px) and spread (px) at the last horizon ($h{=}32$ on A2-div, $16$ on A1 and B).}
\label{tab:e3}
\begin{center}
\small
\setlength{\tabcolsep}{4pt}
\begin{tabular}{lccccccccc}
\toprule
 & \multicolumn{3}{c}{A2-div, $h{=}32$} & \multicolumn{3}{c}{A1 Bouncing Shapes, $h{=}16$} & \multicolumn{3}{c}{B Push-T, $h{=}16$} \\
\cmidrule(lr){2-4}\cmidrule(lr){5-7}\cmidrule(lr){8-10}
Method & manifold & best-of-16 & spread & manifold & best-of-16 & spread & manifold & best-of-16 & spread \\
\midrule
Regression & 1.80 & 32.2 & 0 & 2.04 & 34.0 & 0 & 3.21 & 102.2 & 0 \\
FM-50 & 6.16 & 9.8 & 25.7 & 2.95 & 28.0 & 25.1 & 4.60 & \best{67.4} & 36.4 \\
FM-2 & 6.36 & 10.6 & 26.8 & 3.03 & 23.4 & 26.1 & 4.37 & 69.2 & 29.7 \\
iMF & 8.04 & 10.1 & 29.0 & 3.57 & 33.4 & -- & 5.81 & 84.1 & 57.0 \\
\imlewm{}, $m{=}4$ & 1.74 & 10.5 & 23.9 & \best{1.98} & \best{19.0} & -- & \best{3.20} & 88.7 & 9.2 \\
\imlewm{}, $m{=}16$ & \best{1.68} & \best{9.3} & 27.2 & 2.08 & 19.4 & 17.9 & 3.28 & 79.1 & 21.8 \\
\bottomrule
\end{tabular}
\end{center}
\end{table}

\subsection{Stochastic Push-T: bootstrap intervals}
Table~\ref{tab:bci} gives the paired differences behind Section~\ref{sec:b} and Section~\ref{sec:seq}.

\begin{table}[h]
\caption{Paired differences on the stochastic Push-T multi-future test set (px; negative favors the first model) with 95\% bootstrap intervals over the 128 starting states after averaging seeds, and hierarchical intervals that also resample the three seeds (10{,}000 resamples).}
\label{tab:bci}
\begin{center}
\small
\begin{tabular}{llcccc}
\toprule
Comparison & Metric & Difference & Start-state CI & Hierarchical CI \\
\midrule
\imlewm{} 16 vs.\ FM-50 & $h{=}8$ recall & $-1.8$ & $[-2.5, -1.1]$ & $[-5.2, +2.3]$ \\
 & $h{=}8$ precision & $-5.2$ & $[-5.7, -4.8]$ & $[-8.0, -2.5]$ \\
 & $h{=}16$ recall & $-4.2$ & $[-5.3, -3.1]$ & $[-8.8, +1.1]$ \\
 & $h{=}16$ precision & $-6.6$ & $[-7.6, -5.7]$ & $[-8.5, -4.7]$ \\
\imlewm{} 16 vs.\ FM-2 & $h{=}16$ recall & $-3.7$ & $[-4.9, -2.5]$ & $[-8.7, +1.9]$ \\
 & $h{=}16$ precision & $-5.6$ & $[-6.6, -4.6]$ & $[-7.7, -3.2]$ \\
\imlewm{} 16 vs.\ Shortcut & $h{=}16$ recall & $-8.2$ & $[-10.0, -6.3]$ & $[-11.3, -4.0]$ \\
 & $h{=}16$ precision & $-13.2$ & $[-15.1, -11.3]$ & $[-16.3, -10.2]$ \\
Per-step vs.\ one-step matching (8k) & $h{=}8$ recall & $-7.1$ & $[-8.2, -6.0]$ & $[-9.2, -4.8]$ \\
 & $h{=}16$ recall & $-14.3$ & $[-16.3, -12.5]$ & $[-19.1, -9.9]$ \\
 & $h{=}16$ precision & $-17.0$ & $[-18.4, -15.6]$ & $[-20.4, -14.5]$ \\
\bottomrule
\end{tabular}
\end{center}
\end{table}

\subsection{Energy-score follow-ups}
\label{app:es}
\begin{figure}[h]
\centering
\includegraphics[width=0.9\linewidth]{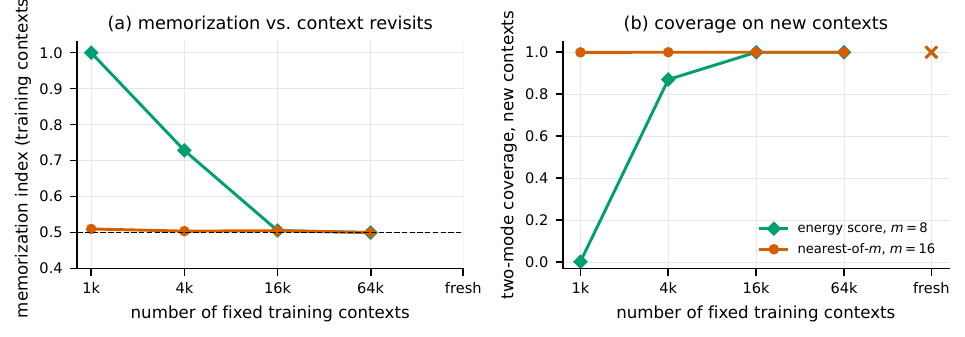}
\caption{Toy two-branch problem with a nuisance appearance vector (two seeds, $m{=}8$ for the energy score and $16$ for \nom{}). (a) Memorization index on training contexts against the number of fixed training contexts. (b) Two-mode coverage on new contexts; ``fresh'' draws a new context at every step.}
\label{fig:toy}
\end{figure}

\begin{table}[h]
\caption{Energy score against \nom{} on bounce (12k steps; two seeds unless noted). A2-div: randomized appearance, contexts never repeat. A2-fix: fixed appearance, no identifying key (same recipe for all rows). The one-step manifold distance is in units of the median real-to-real distance (A2-fix: relative to \imlewm{} $m{=}16$). A1 and the readout comparison are in Appendix~\ref{app:readout}.}
\label{tab:es}
\begin{center}
\footnotesize
\setlength{\tabcolsep}{4pt}
\begin{tabular}{llcccccc}
\toprule
Setting & Method & Coverage & Precision & kNN recall & Memo.\ index & Det.\ err.\ (px) & Manifold $h{=}1$ \\
\midrule
A2-div & Energy score, $m{=}8$ (3 seeds) & 0.36 & 0.61 & 0.77 & 0.87 & 1.06 & 1.19 \\
 & Energy score, $m{=}16$ & 0.28 & 0.61 & 0.60 & 0.97 & 1.06 & -- \\
 & Per-token energy score, $m{=}8$ & 0.22 & 0.55 & 0.44 & 0.97 & 1.33 & 1.12 \\
 & \imlewm{}, $m{=}4$ (3 seeds) & 0.46 & 0.68 & 0.65 & 0.75 & 1.14 & 0.84 \\
 & \imlewm{}, $m{=}16$ (3 seeds) & \best{0.93} & \best{0.82} & \best{0.94} & 0.29 & 1.55 & 0.86 \\
\midrule
A2-fix & Energy score, $m{=}8$ & 0.95 & 0.81 & \best{1.00} & 0.27 & 0.20 & $1.59\times$ \\
 & \imlewm{}, $m{=}4$ & 0.97 & 0.91 & 0.68 & 0.26 & 0.15 & $0.94\times$ \\
 & \imlewm{}, $m{=}16$ & \best{0.99} & \best{0.95} & 0.85 & 0.25 & 0.91 & $1\times$ \\
\bottomrule
\end{tabular}
\end{center}
\end{table}
Figure~\ref{fig:toy} shows the toy experiment of Section~\ref{sec:es}. The task is the two-branch problem of \citet{makabe2026esp} ($u \sim U[0.05, 1]$, $y = \pm\sqrt{u} + 0.08\varepsilon$, equiprobable modes) with a 16-dimensional nuisance vector $a \sim \mathcal{N}(0, I)$ appended to the context, a three-layer MLP generator of width 256 with a 4-dimensional noise input, Adam at $2\times10^{-4}$, batch 256, 15k steps, and the same energy-score estimator as the main runs. With fresh contexts at every step the energy score matches $0.940$ / $0.946$ of its samples to a mode center within $0.24$ ($m = 8$ / $16$) and \nom{} $0.916$ / $0.924$. With 1{,}024 fixed training contexts (about 3{,}750 revisits each) the energy score memorizes (index $0.999$) and covers both modes on $0.2\%$ of new contexts, while \nom{} with $m = 8$ / $16$ is at $0.52$ / $0.51$ and covers $0.99$ / $1.00$. With 4{,}096, 16{,}384 and 65{,}536 contexts the energy score's memorization index is $0.73$, $0.51$ and $0.50$; without the nuisance vector (context $= u$ only) it is $0.50$ even with 1{,}024 contexts. The thresholds are those of this small network and do not transfer; the 8M model on A2-div saw each of its $11{,}559$ stochastic training contexts about 38 times and memorized.

The follow-up runs on A2-fix and with the per-token score were preceded by five recorded predictions, all of which held (two seeds each): (P1) the energy score with $m{=}8$ does not memorize on A2-fix (index $\le 0.40$; observed $0.268$); (P2) its coverage on A2-fix exceeds its A2-div value but stays below \imlewm{} with $m{=}16$ in both coverage and precision (observed $0.947$ vs.\ $0.987$ and $0.814$ vs.\ $0.954$); (P3) \imlewm{} with $m{=}4$ does not memorize on A2-fix ($\le 0.35$; observed $0.256$); (P4) the per-token energy score memorizes on A2-div (index $\ge 0.7$, coverage $\le 0.5$; observed $0.970$ and $0.224$); (P5) its one-step manifold distance exceeds $1.1$ (observed $1.12$; \imlewm{} $0.86$). On A2-fix the energy score's kNN recall is $1.00$ against $0.85$ for \imlewm{}, so its samples spread more widely than the outcomes; its deterministic-transition error is the lowest ($0.20$\,px). On A2-div its learning curves peak in coverage at 6k steps ($0.50$) and decline as the index rises ($0.42 \to 0.87$); the two $m{=}16$ seeds reach an index of $0.84$ and $0.87$ at 8k steps against $0.51$--$0.72$ for the three $m{=}8$ seeds. Token-space diagnostics at $h{=}1$ on A2-div (three seeds, evaluation contexts): spread-to-error ratio $0.80$ (energy score) vs.\ $0.64$ (\imlewm{} 16), $0.38$ (\imlewm{} 4), $1.15$ (FM-50); best-of-16 token error $0.468$ vs.\ $0.271$; mean-prediction error $0.451$ vs.\ $0.396$; readout spread $1.7$ vs.\ $3.9$\,px; one-step manifold distance $1.19$ vs.\ $0.86$.

\subsection{Per-step matching}
\label{app:seq}

\begin{table}[h]
\caption{Fed-back-state ablation on Bouncing Shapes (8M, $m{=}16$, 8k steps, two seeds): manifold distance, best-of-16 readout error and spread of the 16 rollouts at $h{=}16$ (the true futures spread by $23.6$\,px), and F1 error. All three predictions recorded before the ablation held (S1--S3 below).}
\label{tab:seqabl}
\begin{center}
\small
\begin{tabular}{lcccc}
\toprule
Fed-back state & Manifold & Best-of-16 & Spread & F1 \\
\midrule
One-step matching & 2.17 & 23.2 & 17.0 & 8.74 \\
Per-step, feed matched & 1.92 & \best{17.7} & 26.3 & \best{8.51} \\
Per-step, feed random candidate & \best{1.81} & 22.9 & 12.4 & 8.77 \\
Per-step, feed true frame & 2.14 & 23.6 & 16.6 & 8.62 \\
\bottomrule
\end{tabular}
\end{center}
\end{table}

\begin{figure}[h]
\centering
\includegraphics[width=\linewidth]{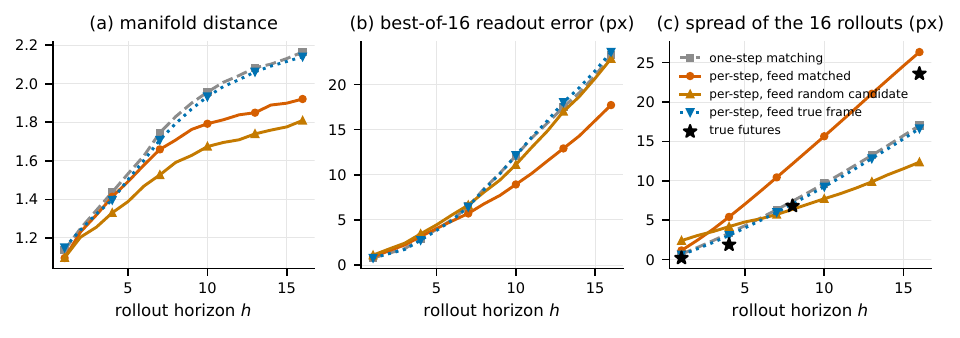}
\caption{The fed-back-state ablation on Bouncing Shapes (two seeds). (a) Any self-generated context, matched or random, keeps the rollout on the manifold; the true frame does not. (b) Only the candidate nearest the truth improves the best-of-16 error. (c) The matched variant is the only one whose 16 rollouts spread like the true futures at $h{=}16$; the random variant jitters at every step but does not form distinct futures, and spreads to half the true spread.}
\label{fig:seqabl}
\end{figure}
Per-seed Push-T recall errors at $h{=}16$ (one-step $\to$ per-step): $43.7 \to 30.3$, $42.6 \to 32.2$, $44.7 \to 25.5$. On the deterministic Push-T control (one seed) the best-of-16 T-vertex error at $h = 4 / 8 / 16$ moves from $35.8 / 60.4 / 72.3$ to $29.2 / 45.6 / 58.9$\,px and the manifold distance at $h{=}16$ from $3.90$ to $3.64$. On A2-div (two seeds, 8k steps) the manifold distance at $h = 8 / 16 / 32$ moves from $1.31 / 1.47 / 1.83$ to $1.19 / 1.26 / 1.40$ while one-step coverage and precision move from $0.92 / 0.77$ to $0.90 / 0.73$ and the deterministic-transition error from $2.91$ to $3.97$\,px; the best-of-16 error is unchanged. On A1 (two seeds) F1 moves from $8.74$ to $8.51$ (precision $9.26 \to 9.77$, recall $8.28 \to 7.55$), the manifold distance at $h{=}16$ from $2.17$ to $1.92$ and the best-of-16 error from $23.2$ to $17.7$\,px. The fed-back-state ablation was preceded by three recorded predictions: (S1) feeding back the true frame yields less than half of the manifold gain (manifold at $h{=}16$ $\ge 2.05$; observed $2.14$); (S2) feeding back a random candidate yields more than half ($\le 2.05$; observed $1.81$); (S3) the random candidate's best-of-16 error exceeds the matched one's by at least $1$\,px ($\ge 18.7$; observed $22.9$). The spread of the 16 rollouts at $h = 1 / 4 / 8 / 16$ is $0.20 / 1.90 / 6.81 / 23.6$\,px for the true futures, $0.70 / 3.35 / 7.42 / 17.0$ for one-step matching, $1.16 / 5.42 / 12.2 / 26.3$ for the matched variant, $2.41 / 4.19 / 6.38 / 12.4$ for the random variant and $0.61 / 3.07 / 7.08 / 16.6$ for the true-frame variant. Training times: one-step $0.68$\,h, per-step variants $2.7$\,h (8M, 8k steps).

\subsection{Planning details}
\label{app:plan}
\begin{table}[h]
\caption{Planning with CEM-MPC. Left: stochastic Push-T, 100 episodes $\times$ 3 seeds (FM-50: 10 episodes, 1 seed), planning time per MPC step. Right: branch-risk task (push along the pit edge), 100 episodes $\times$ 3 seeds, and the model-estimated pit probability after one push on the ice (true $0.33$) and after five straight pushes from 200 start states (true $0.43$).}
\label{tab:plan}
\begin{center}
\scriptsize
\setlength{\tabcolsep}{4pt}
\begin{tabular}{lcc}
\toprule
Stochastic Push-T & Success & s / step \\
\midrule
Regression & \ms{31.7}{3.3}\% & 4.3 \\
FM-2 & \ms{34.7}{3.1}\% & 8.5 \\
FM-50 & 30\% & 200 \\
\imlewm{}, $m{=}16$ & \ms{31.7}{2.1}\% & \best{4.0} \\
\bottomrule
\end{tabular}
\hfill
\begin{tabular}{lcccc}
\toprule
Branch-risk task & Success & Pit & One-push pit prob.\ (true 0.33) & Five-push pit prob.\ (true 0.43) \\
\midrule
Oracle (true simulator) & 76\% & 7\% & 0.33 & 0.43 \\
Regression & \ms{56.3}{3.4}\% & \ms{24.3}{2.5}\% & \ms{0.61}{0.10} & \ms{0.26}{0.19} \\
FM-2 & \best{\ms{67.0}{1.4}\%} & \best{\ms{18.7}{0.9}\%} & \ms{0.14}{0.00} & \ms{0.38}{0.06} \\
FM-50 (estimates only) & -- & -- & \ms{0.31}{0.02} & 0.44 \\
\imlewm{}, $m{=}16$ & \ms{55.3}{2.6}\% & \ms{31.0}{2.2}\% & \best{\ms{0.36}{0.03}} & \best{\ms{0.44}{0.04}} \\
\bottomrule
\end{tabular}
\end{center}
\end{table}
On stochastic Push-T the final block position and angle errors are $46.6$\,px / $42.0^\circ$ (regression), $40.6$ / $35.2$ (FM-2), $45.2$ / $20.0$ (FM-50, 10 episodes) and $49.5$ / $44.9$ (\imlewm{}). The one-step contact error (block displacement after a contact transition, true $15.9$\,px) is $11.7$ / $11.2$\,px for regression at 12k / 36k steps, $12.4$ / $12.6$ for FM-2 / FM-50 and $12.2$ / $12.3$ for \imlewm{} with $m = 4$ / $16$ (seed 0). On the branch-risk task the per-start absolute error of the five-push pit probability is $0.078$ (\imlewm{}), $0.086$ (FM-50), $0.096$ (FM-2) and $0.462$ (regression), and the timeout rates are $13.7\%$, $14.3\%$, $19.3\%$ and $17\%$ (oracle). Wall-clock for 100 episodes: oracle $100$\,s, regression $523$\,s ($m_p{=}1$), \imlewm{} $3{,}633$\,s, FM-2 $9{,}067$\,s.

\section{Negative results}
\label{app:negative}
The pilot study tested four additions proposed in our original design; none survived.

\paragraph{Trajectory-level matching.}
Matching $m{=}4$ candidate rollouts of length $H{=}8$ as whole trajectories (with a curriculum $H = 1 \to 4 \to 8$) on the push environment was worse than one-step matching at every horizon: best-of-16 position error $1.31$ vs.\ $0.35$\,px at $h{=}1$, $3.89$ vs.\ $2.19$ at $h{=}8$, $4.67$ vs.\ $3.03$ at $h{=}16$ (paired bootstrap intervals exclude zero), one-step coverage $0.71$ vs.\ $0.98$, and $3.3\times$ the training time; only the error of the sample mean improved ($9.18$ vs.\ $9.94$ at $h{=}16$), the signature of averaging. With two to three modes per step an 8-step trajectory has hundreds of trajectory modes, far beyond $m{=}4$, so the matched candidate is the one close to all of them on average. Per-step matching (Section~\ref{sec:seqmatch}) replaced it.

\paragraph{Multi-stage supervision.}
On the fixed $16\times16$ token grid, supervising intermediate heads with spatially pooled targets ($4\times4$, $8\times8$, $16\times16$) or with PCA-truncated targets (32, 128, 384 components) gave the same coverage as a single head within the bootstrap interval ($0.960$ and $0.959$ vs.\ $0.967$ at $K{=}4$) and the same precision ($0.916$, $0.922$ vs.\ $0.926$). The final recipe uses a single head.

\paragraph{Variance-penalized planning.}
A CEM cost of mean plus $\lambda$ times the spread of the imagined goal distances lowered the success rate of the branch-risk task even when the true simulator served as the world model: $78\%$, $77\%$, $70\%$ and $50\%$ at $\lambda = 0, 0.5, 1, 2$, with the pit rate rising to $35\%$. A certain failure has zero variance, so the penalty prefers it to an uncertain success. With \imlewm{} as the model (two seeds, 60 episodes each) the pooled success was $62.5\%$ at $\lambda = 0$ and $61.7\%$ at $\lambda = 0.5$.

\paragraph{Counterfactual action contrast.}
We had anticipated that the noise $z$ might absorb the effect of the action. It does not: on the push environment the counterfactual precision (the fraction of samples under a random other action that stay away from the true outcome) and the action sensitivity (the change in output per change in action, relative to the true change) are $0.95$ and $0.96$ with the action on its own FiLM path, $0.92$ and $0.97$ with the action merged into the noise path, and $0.93$ and $0.975$ with an added margin term, against $0.83$ and $0.94$ for regression. There was no shortcut to repair.

\paragraph{Low-dimensional matching and larger budgets on BAIR.}
See Table~\ref{tab:bair}: $m{=}64$ raises the spread-to-error ratio from $0.44$--$0.51$ to $0.53$--$0.55$ at $2.4\times$ the training time, and matching in a 512-dimensional pooled space lowers it to $0.31$--$0.38$; neither approaches FM-50 on distribution metrics.

\section{Statistical methodology}
\label{app:stats}
Per-context intervals are percentile bootstrap intervals (2{,}000 resamples) over the 400 evaluation contexts of a run, used in the pilot and recipe tables. Seed statistics are means and standard deviations over the listed seeds. For the Push-T multi-future test set, each model's per-start errors were stored; the start-state interval averages the seeds first and resamples the 128 starts, and the hierarchical interval resamples the seeds with replacement and then the starts, 10{,}000 times each. Per-step and one-step matching runs with the same seed are paired. All comparisons report the direction and the magnitude of the difference; where an interval includes zero we say so in the text.


\begin{thebibliography}{81}
\providecommand{\natexlab}[1]{#1}
\providecommand{\url}[1]{\texttt{#1}}
\expandafter\ifx\csname urlstyle\endcsname\relax
  \providecommand{\doi}[1]{doi: #1}\else
  \providecommand{\doi}{doi: \begingroup \urlstyle{rm}\Url}\fi

\bibitem[Aghabozorgi et~al.(2023)Aghabozorgi, Peng, and
  Li]{aghabozorgi2023adaptive}
Mehran Aghabozorgi, Shichong Peng, and Ke~Li.
\newblock Adaptive {IMLE} for few-shot pretraining-free generative modelling.
\newblock In \emph{International Conference on Machine Learning (ICML)}, 2023.

\bibitem[Aghabozorgi et~al.(2026)Aghabozorgi, Moazeni, Zhang, and
  Li]{aghabozorgi2026wimle}
Mehran Aghabozorgi, Alireza Moazeni, Yanshu Zhang, and Ke~Li.
\newblock {WIMLE}: Uncertainty-aware world models with {IMLE} for
  sample-efficient continuous control.
\newblock In \emph{International Conference on Learning Representations
  (ICLR)}, 2026.

\bibitem[{AgiBot Research Team} et~al.(2026){AgiBot Research Team}, Liu, Zhao,
  Yang, Chen, et~al.]{agibot2026geact}
{AgiBot Research Team}, Renhang Liu, Wenzhi Zhao, Zhuo Yang, Liliang Chen,
  et~al.
\newblock {GE-Act 2.0}: Pretraining and scaling a world-action model for
  robotic manipulation.
\newblock \emph{arXiv preprint arXiv:2609.05588}, 2026.

\bibitem[Assran et~al.(2025)Assran, Bardes, Fan, Garrido, Howes, Komeili,
  Muckley, Rizvi, Roberts, Sinha, Zholus, Arnaud, Gejji, Martin, Hogan, Dugas,
  Bojanowski, Khalidov, Labatut, Massa, Szafraniec, Krishnakumar, Li, Ma,
  Chandar, Meier, LeCun, Rabbat, and Ballas]{assran2025vjepa2}
Mido Assran, Adrien Bardes, David Fan, Quentin Garrido, Russell Howes, Mojtaba
  Komeili, Matthew Muckley, Ammar Rizvi, Claire Roberts, Koustuv Sinha, Artem
  Zholus, Sergio Arnaud, Abha Gejji, Ada Martin, Francois~Robert Hogan, Daniel
  Dugas, Piotr Bojanowski, Vasil Khalidov, Patrick Labatut, Francisco Massa,
  Marc Szafraniec, Kapil Krishnakumar, Yong Li, Xiaodong Ma, Sarath Chandar,
  Franziska Meier, Yann LeCun, Michael Rabbat, and Nicolas Ballas.
\newblock {V-JEPA 2}: Self-supervised video models enable understanding,
  prediction and planning.
\newblock \emph{arXiv preprint arXiv:2506.09985}, 2025.

\bibitem[Baldassarre et~al.(2025)Baldassarre, Szafraniec, Terver, Khalidov,
  Massa, LeCun, Labatut, Seitzer, and Bojanowski]{baldassarre2025dinoworld}
Federico Baldassarre, Marc Szafraniec, Basile Terver, Vasil Khalidov, Francisco
  Massa, Yann LeCun, Patrick Labatut, Maximilian Seitzer, and Piotr Bojanowski.
\newblock Back to the features: {DINO} as a foundation for video world models.
\newblock \emph{arXiv preprint arXiv:2507.19468}, 2025.

\bibitem[Bengio et~al.(2015)Bengio, Vinyals, Jaitly, and
  Shazeer]{bengio2015scheduled}
Samy Bengio, Oriol Vinyals, Navdeep Jaitly, and Noam Shazeer.
\newblock Scheduled sampling for sequence prediction with recurrent neural
  networks.
\newblock In \emph{Advances in Neural Information Processing Systems
  (NeurIPS)}, 2015.

\bibitem[Bennett(2026)]{bennett2026freedom}
Michael~Timothy Bennett.
\newblock Why the third axis is freedom.
\newblock \emph{arXiv preprint arXiv:2608.05423}, 2026.

\bibitem[Bhattacharyya et~al.(2018)Bhattacharyya, Schiele, and
  Fritz]{bhattacharyya2018bestofmany}
Apratim Bhattacharyya, Bernt Schiele, and Mario Fritz.
\newblock Accurate and diverse sampling of sequences based on a ``best of
  many'' sample objective.
\newblock In \emph{IEEE/CVF Conference on Computer Vision and Pattern
  Recognition (CVPR)}, 2018.

\bibitem[Boduljak et~al.(2025)Boduljak, Lan, Rupprecht, and
  Vedaldi]{boduljak2025vfmf}
Gabrijel Boduljak, Yushi Lan, Christian Rupprecht, and Andrea Vedaldi.
\newblock {VFMF}: World modeling by forecasting vision foundation model
  features.
\newblock \emph{arXiv preprint arXiv:2512.11225}, 2025.

\bibitem[Carlini et~al.(2023)Carlini, Hayes, Nasr, Jagielski, Sehwag,
  Tram{\`e}r, Balle, Ippolito, and Wallace]{carlini2023extracting}
Nicholas Carlini, Jamie Hayes, Milad Nasr, Matthew Jagielski, Vikash Sehwag,
  Florian Tram{\`e}r, Borja Balle, Daphne Ippolito, and Eric Wallace.
\newblock Extracting training data from diffusion models.
\newblock In \emph{USENIX Security Symposium}, 2023.

\bibitem[Chakraborty \& Zhang(2021)Chakraborty and
  Zhang]{chakraborty2021framework}
Shubhadeep Chakraborty and Xianyang Zhang.
\newblock A new framework for distance and kernel-based metrics in high
  dimensions.
\newblock \emph{Electronic Journal of Statistics}, 15\penalty0 (2):\penalty0
  5455--5522, 2021.

\bibitem[Chen et~al.(2024)Chen, Mart{\'i}~Mons{\'o}, Du, Simchowitz, Tedrake,
  and Sitzmann]{chen2024diffusionforcing}
Boyuan Chen, Diego Mart{\'i}~Mons{\'o}, Yilun Du, Max Simchowitz, Russ Tedrake,
  and Vincent Sitzmann.
\newblock Diffusion forcing: Next-token prediction meets full-sequence
  diffusion.
\newblock In \emph{Advances in Neural Information Processing Systems
  (NeurIPS)}, 2024.

\bibitem[Cheung \& He(2026)Cheung and He]{cheung2026fidelity}
Joshua Cheung and Zhang He.
\newblock One-step prediction fidelity is a poor proxy for planning performance
  in {JEPA} world models.
\newblock \emph{arXiv preprint arXiv:2610.19375}, 2026.

\bibitem[Cort{\'e}s et~al.(2025)Cort{\'e}s, Rehm, and
  Letzelter]{cortes2025timemcl}
Adrien Cort{\'e}s, R{\'e}mi Rehm, and Victor Letzelter.
\newblock Winner-takes-all for multivariate probabilistic time series
  forecasting.
\newblock In \emph{International Conference on Machine Learning (ICML)}, 2025.

\bibitem[Cover \& Thomas(2006)Cover and Thomas]{cover2006elements}
Thomas~M. Cover and Joy~A. Thomas.
\newblock \emph{Elements of Information Theory}.
\newblock Wiley, 2nd edition, 2006.

\bibitem[Dawid(2007)]{dawid2007geometry}
A.~Philip Dawid.
\newblock The geometry of proper scoring rules.
\newblock \emph{Annals of the Institute of Statistical Mathematics},
  59:\penalty0 77--93, 2007.

\bibitem[De~Bortoli et~al.(2025)De~Bortoli, Galashov, Guntupalli, Zhou, Murphy,
  Gretton, and Doucet]{debortoli2025distributional}
Valentin De~Bortoli, Alexandre Galashov, J.~Swaroop Guntupalli, Guangyao Zhou,
  Kevin Murphy, Arthur Gretton, and Arnaud Doucet.
\newblock Distributional diffusion models with scoring rules.
\newblock In \emph{International Conference on Machine Learning (ICML)}, 2025.

\bibitem[Delattre et~al.(2025)Delattre, Delattre, V{\'e}rine, and
  Allauzen]{delattre2025cdq}
Blaise Delattre, Sylvain Delattre, Alexandre V{\'e}rine, and Alexandre
  Allauzen.
\newblock Conditional distribution quantization in machine learning.
\newblock \emph{arXiv preprint arXiv:2502.07151}, 2025.

\bibitem[Ebert et~al.(2017)Ebert, Finn, Lee, and Levine]{ebert2017bair}
Frederik Ebert, Chelsea Finn, Alex~X. Lee, and Sergey Levine.
\newblock Self-supervised visual planning with temporal skip connections.
\newblock In \emph{Conference on Robot Learning (CoRL)}, 2017.

\bibitem[Fan et~al.(2017)Fan, Su, and Guibas]{fan2017pointset}
Haoqiang Fan, Hao Su, and Leonidas~J. Guibas.
\newblock A point set generation network for {3D} object reconstruction from a
  single image.
\newblock In \emph{IEEE Conference on Computer Vision and Pattern Recognition
  (CVPR)}, 2017.

\bibitem[Frans et~al.(2025)Frans, Hafner, Levine, and
  Abbeel]{frans2025shortcut}
Kevin Frans, Danijar Hafner, Sergey Levine, and Pieter Abbeel.
\newblock One step diffusion via shortcut models.
\newblock In \emph{International Conference on Learning Representations
  (ICLR)}, 2025.

\bibitem[Gao et~al.(2026)Gao, Guo, Guan, Huang, Ma, Xiao, Xiong, and
  Wen]{gao2026sword}
Jiaxuan Gao, Yongjian Guo, Zhong Guan, Wen Huang, Wanlun Ma, Xi~Xiao, Junwu
  Xiong, and Sheng Wen.
\newblock Sword: Style-robust world models as simulators via dynamic latent
  bootstrapping for {VLA} policy post-training.
\newblock \emph{arXiv preprint arXiv:2605.07288}, 2026.

\bibitem[Geman \& McClure(1987)Geman and McClure]{geman1987statistical}
Stuart Geman and Donald~E. McClure.
\newblock Statistical methods for tomographic image reconstruction.
\newblock \emph{Bulletin of the International Statistical Institute},
  52\penalty0 (4):\penalty0 5--21, 1987.

\bibitem[Geng et~al.(2025{\natexlab{a}})Geng, Deng, Bai, Kolter, and
  He]{geng2025meanflow}
Zhengyang Geng, Mingyang Deng, Xingjian Bai, J.~Zico Kolter, and Kaiming He.
\newblock Mean flows for one-step generative modeling.
\newblock In \emph{Advances in Neural Information Processing Systems
  (NeurIPS)}, 2025{\natexlab{a}}.

\bibitem[Geng et~al.(2025{\natexlab{b}})Geng, Lu, Wu, Shechtman, Kolter, and
  He]{geng2025imf}
Zhengyang Geng, Yiyang Lu, Zongze Wu, Eli Shechtman, J.~Zico Kolter, and
  Kaiming He.
\newblock Improved mean flows: On the challenges of fastforward generative
  models.
\newblock \emph{arXiv preprint arXiv:2512.02012}, 2025{\natexlab{b}}.

\bibitem[Gladstone et~al.(2026)Gladstone, Ji, and Du]{gladstone2026xm}
Alexi Gladstone, Heng Ji, and Yilun Du.
\newblock Explorative modeling: Unlocking a third pretraining axis and
  end-to-end generation.
\newblock \emph{arXiv preprint arXiv:2607.27372}, 2026.

\bibitem[Gneiting \& Raftery(2007)Gneiting and Raftery]{gneiting2007strictly}
Tilmann Gneiting and Adrian~E. Raftery.
\newblock Strictly proper scoring rules, prediction, and estimation.
\newblock \emph{Journal of the American Statistical Association}, 102\penalty0
  (477):\penalty0 359--378, 2007.

\bibitem[Gu et~al.(2025)Gu, Du, Pang, Li, Lin, and Wang]{gu2025memorization}
Xiangming Gu, Chao Du, Tianyu Pang, Chongxuan Li, Min Lin, and Ye~Wang.
\newblock On memorization in diffusion models.
\newblock \emph{Transactions on Machine Learning Research}, 2025.

\bibitem[Guzman-Rivera et~al.(2012)Guzman-Rivera, Batra, and
  Kohli]{guzman2012mcl}
Abner Guzman-Rivera, Dhruv Batra, and Pushmeet Kohli.
\newblock Multiple choice learning: Learning to produce multiple structured
  outputs.
\newblock In \emph{Advances in Neural Information Processing Systems
  (NeurIPS)}, 2012.

\bibitem[Hafner et~al.(2019)Hafner, Lillicrap, Fischer, Villegas, Ha, Lee, and
  Davidson]{hafner2019planet}
Danijar Hafner, Timothy Lillicrap, Ian Fischer, Ruben Villegas, David Ha,
  Honglak Lee, and James Davidson.
\newblock Learning latent dynamics for planning from pixels.
\newblock In \emph{International Conference on Machine Learning (ICML)}, 2019.

\bibitem[Hafner et~al.(2025)Hafner, Yan, and Lillicrap]{hafner2025dreamer4}
Danijar Hafner, Wilson Yan, and Timothy Lillicrap.
\newblock Training agents inside of scalable world models.
\newblock \emph{arXiv preprint arXiv:2509.24527}, 2025.

\bibitem[Hosseinkhani et~al.(2026)Hosseinkhani, Peng, Shramko, Aghabozorgi,
  Qian, Engst, Moazeni, Jayaraman, and Li]{hosseinkhani2026imlevla}
Kian Hosseinkhani, Qinhe Peng, George Shramko, Mehran Aghabozorgi, Jianing
  Qian, Tristan Engst, Alireza Moazeni, Dinesh Jayaraman, and Ke~Li.
\newblock {IMLE-VLA}: Fast single-step action generation for
  vision-language-action policies.
\newblock In \emph{IEEE/RSJ International Conference on Intelligent Robots and
  Systems (IROS)}, 2026.

\bibitem[Hsu et~al.(2026)Hsu, Gwak, Lim, Li, and Erichson]{hsu2026vsf}
Hans Hao-Hsun Hsu, Minseon Gwak, Soon~Hoe Lim, Pan Li, and N.~Benjamin
  Erichson.
\newblock Variational streaming flow: Probabilistic forecasting in physical
  time.
\newblock \emph{arXiv preprint arXiv:2610.00976}, 2026.

\bibitem[Huang et~al.(2025)Huang, Li, He, Zhou, and
  Shechtman]{huang2025selfforcing}
Xun Huang, Zhengqi Li, Guande He, Mingyuan Zhou, and Eli Shechtman.
\newblock Self forcing: Bridging the train-test gap in autoregressive video
  diffusion.
\newblock In \emph{Advances in Neural Information Processing Systems
  (NeurIPS)}, 2025.

\bibitem[Karypidis et~al.(2026)Karypidis, Gidaris, and
  Komodakis]{karypidis2026latentforesight}
Efstathios Karypidis, Spyros Gidaris, and Nikos Komodakis.
\newblock {Latent-Foresight}: End-to-end learning predictable representations
  for latent world models.
\newblock \emph{arXiv preprint arXiv:2610.01942}, 2026.

\bibitem[Kim \& Lee(2026)Kim and Lee]{kim2026memorization}
Gwangho Kim and Sungyoon Lee.
\newblock Memorization and malign generalization in conditional diffusion
  models with random features.
\newblock \emph{arXiv preprint arXiv:2610.11288}, 2026.

\bibitem[Kynk{\"a}{\"a}nniemi et~al.(2019)Kynk{\"a}{\"a}nniemi, Karras, Laine,
  Lehtinen, and Aila]{kynkaanniemi2019improved}
Tuomas Kynk{\"a}{\"a}nniemi, Tero Karras, Samuli Laine, Jaakko Lehtinen, and
  Timo Aila.
\newblock Improved precision and recall metric for assessing generative models.
\newblock In \emph{Advances in Neural Information Processing Systems
  (NeurIPS)}, 2019.

\bibitem[Lamb et~al.(2016)Lamb, Goyal, Zhang, Zhang, Courville, and
  Bengio]{lamb2016professor}
Alex Lamb, Anirudh Goyal, Ying Zhang, Saizheng Zhang, Aaron Courville, and
  Yoshua Bengio.
\newblock Professor forcing: A new algorithm for training recurrent networks.
\newblock In \emph{Advances in Neural Information Processing Systems
  (NeurIPS)}, 2016.

\bibitem[Lee et~al.(2026{\natexlab{a}})Lee, Agarwal, and Suh]{lee2026beckmann}
Adam Lee, Shobhit Agarwal, and June~Pyo Suh.
\newblock Beckmann world models: Direct terminal-map learning for one-step
  action-conditioned prediction.
\newblock \emph{arXiv preprint arXiv:2610.26903}, 2026{\natexlab{a}}.

\bibitem[Lee et~al.(2026{\natexlab{b}})Lee, Bui, Zhou, Li, Chen, and
  Li]{lee2026imlempc}
Grayson Lee, Minh Bui, Shuzi Zhou, Yankai Li, Mo~Chen, and Ke~Li.
\newblock Implicit maximum likelihood estimation for real-time generative model
  predictive control.
\newblock In \emph{IEEE International Conference on Robotics and Automation
  (ICRA)}, 2026{\natexlab{b}}.

\bibitem[Lee et~al.(2016)Lee, Purushwalkam, Cogswell, Ranjan, Crandall, and
  Batra]{lee2016smcl}
Stefan Lee, Senthil Purushwalkam, Michael Cogswell, Viresh Ranjan, David
  Crandall, and Dhruv Batra.
\newblock Stochastic multiple choice learning for training diverse deep
  ensembles.
\newblock In \emph{Advances in Neural Information Processing Systems
  (NeurIPS)}, 2016.

\bibitem[Letzelter et~al.(2024)Letzelter, Perera, Rommel, Fontaine, Essid,
  Richard, and P{\'e}rez]{letzelter2024wta}
Victor Letzelter, David Perera, C{\'e}dric Rommel, Mathieu Fontaine, Slim
  Essid, Ga{\"e}l Richard, and Patrick P{\'e}rez.
\newblock Winner-takes-all learners are geometry-aware conditional density
  estimators.
\newblock In \emph{International Conference on Machine Learning (ICML)}, 2024.

\bibitem[Li \& Malik(2018)Li and Malik]{li2018imle}
Ke~Li and Jitendra Malik.
\newblock Implicit maximum likelihood estimation.
\newblock \emph{arXiv preprint arXiv:1809.09087}, 2018.

\bibitem[Li et~al.(2019)Li, Zhang, and Malik]{li2019diverse}
Ke~Li, Tianhao Zhang, and Jitendra Malik.
\newblock Diverse image synthesis from semantic layouts via conditional {IMLE}.
\newblock In \emph{IEEE/CVF International Conference on Computer Vision
  (ICCV)}, 2019.

\bibitem[Li et~al.(2020)Li, Peng, Zhang, and Malik]{li2020multimodal}
Ke~Li, Shichong Peng, Tianhao Zhang, and Jitendra Malik.
\newblock Multimodal image synthesis with conditional implicit maximum
  likelihood estimation.
\newblock \emph{International Journal of Computer Vision}, 128:\penalty0
  2607--2628, 2020.

\bibitem[Lipman et~al.(2023)Lipman, Chen, Ben-Hamu, Nickel, and
  Le]{lipman2023flow}
Yaron Lipman, Ricky T.~Q. Chen, Heli Ben-Hamu, Maximilian Nickel, and Matt Le.
\newblock Flow matching for generative modeling.
\newblock In \emph{International Conference on Learning Representations
  (ICLR)}, 2023.

\bibitem[Liu et~al.(2022)Liu, Mao, Wu, Feichtenhofer, Darrell, and
  Xie]{liu2022convnext}
Zhuang Liu, Hanzi Mao, Chao-Yuan Wu, Christoph Feichtenhofer, Trevor Darrell,
  and Saining Xie.
\newblock A {ConvNet} for the 2020s.
\newblock In \emph{IEEE/CVF Conference on Computer Vision and Pattern
  Recognition (CVPR)}, 2022.

\bibitem[Lu et~al.(2026)Lu, Chen, Ye, and Du]{lu2026driftworld}
Susie Lu, Haonan Chen, Weirui Ye, and Yilun Du.
\newblock {DriftWorld}: Fast world modeling through drifting.
\newblock \emph{arXiv preprint arXiv:2607.15065}, 2026.

\bibitem[Makabe et~al.(2026)Makabe, Kim, and Matsushita]{makabe2026esp}
Lilika Makabe, Heecheol Kim, and Yasuyuki Matsushita.
\newblock {ESP}: Energy-score policy for one-step multimodal action generation.
\newblock \emph{arXiv preprint arXiv:2610.07696}, 2026.

\bibitem[Makansi et~al.(2019)Makansi, Ilg, {\c{C}}i{\c{c}}ek, and
  Brox]{makansi2019ewta}
Osama Makansi, Eddy Ilg, {\"O}zg{\"u}n {\c{C}}i{\c{c}}ek, and Thomas Brox.
\newblock Overcoming limitations of mixture density networks: A sampling and
  fitting framework for multimodal future prediction.
\newblock In \emph{IEEE/CVF Conference on Computer Vision and Pattern
  Recognition (CVPR)}, 2019.

\bibitem[Mao et~al.(2019)Mao, Lee, Tseng, Ma, and Yang]{mao2019msgan}
Qi~Mao, Hsin-Ying Lee, Hung-Yu Tseng, Siwei Ma, and Ming-Hsuan Yang.
\newblock Mode seeking generative adversarial networks for diverse image
  synthesis.
\newblock In \emph{IEEE/CVF Conference on Computer Vision and Pattern
  Recognition (CVPR)}, 2019.

\bibitem[Marcotte et~al.(2023)Marcotte, Zantedeschi, Drouin, and
  Chapados]{marcotte2023regions}
{\'E}tienne Marcotte, Valentina Zantedeschi, Alexandre Drouin, and Nicolas
  Chapados.
\newblock Regions of reliability in the evaluation of multivariate
  probabilistic forecasts.
\newblock In \emph{International Conference on Machine Learning (ICML)}, 2023.

\bibitem[Meehan et~al.(2020)Meehan, Chaudhuri, and
  Dasgupta]{meehan2020nonparametric}
Casey Meehan, Kamalika Chaudhuri, and Sanjoy Dasgupta.
\newblock A non-parametric test to detect data-copying in generative models.
\newblock In \emph{International Conference on Artificial Intelligence and
  Statistics (AISTATS)}, 2020.

\bibitem[Oquab et~al.(2024)Oquab, Darcet, Moutakanni, Vo, Szafraniec, Khalidov,
  Fernandez, Haziza, Massa, El-Nouby, Assran, Ballas, Galuba, Howes, Huang, Li,
  Misra, Rabbat, Sharma, Synnaeve, Xu, J{\'e}gou, Mairal, Labatut, Joulin, and
  Bojanowski]{oquab2024dinov2}
Maxime Oquab, Timoth{\'e}e Darcet, Th{\'e}o Moutakanni, Huy Vo, Marc
  Szafraniec, Vasil Khalidov, Pierre Fernandez, Daniel Haziza, Francisco Massa,
  Alaaeldin El-Nouby, Mahmoud Assran, Nicolas Ballas, Wojciech Galuba, Russell
  Howes, Po-Yao Huang, Shang-Wen Li, Ishan Misra, Michael Rabbat, Vasu Sharma,
  Gabriel Synnaeve, Hu~Xu, Herv{\'e} J{\'e}gou, Julien Mairal, Patrick Labatut,
  Armand Joulin, and Piotr Bojanowski.
\newblock {DINOv2}: Learning robust visual features without supervision.
\newblock \emph{Transactions on Machine Learning Research}, 2024.

\bibitem[Park et~al.(2026)Park, Kim, Du, Chen, and Han]{park2026priming}
Woobin Park, Jaeyeon Kim, Yilun Du, Sitan Chen, and Bohyung Han.
\newblock Are we overthinking one-step generation?
\newblock \emph{arXiv preprint arXiv:2610.20331}, 2026.

\bibitem[Peng et~al.(2022)Peng, Moazeni, and Li]{peng2022chimle}
Shichong Peng, Alireza Moazeni, and Ke~Li.
\newblock {CHIMLE}: Conditional hierarchical {IMLE} for multimodal conditional
  image synthesis.
\newblock In \emph{Advances in Neural Information Processing Systems
  (NeurIPS)}, 2022.

\bibitem[Perera et~al.(2024)Perera, Letzelter, Mariotte, Cort{\'e}s, Chen,
  Essid, and Richard]{perera2024amcl}
David Perera, Victor Letzelter, Th{\'e}o Mariotte, Adrien Cort{\'e}s, Mickael
  Chen, Slim Essid, and Ga{\"e}l Richard.
\newblock Annealed multiple choice learning: Overcoming limitations of
  winner-takes-all with annealing.
\newblock In \emph{Advances in Neural Information Processing Systems
  (NeurIPS)}, 2024.

\bibitem[Pinson \& Tastu(2013)Pinson and Tastu]{pinson2013discrimination}
Pierre Pinson and Julija Tastu.
\newblock Discrimination ability of the energy score.
\newblock Technical Report Technical Report 2013-15, Technical University of
  Denmark, DTU Compute, 2013.

\bibitem[Porcher et~al.(2026)Porcher, Carion, Alahari, and
  Chen]{porcher2026flowwm}
Fran{\c{c}}ois Porcher, Nicolas Carion, Karteek Alahari, and Shizhe Chen.
\newblock Flow matching in feature space for stochastic world modeling.
\newblock \emph{arXiv preprint arXiv:2606.29059}, 2026.

\bibitem[Rana et~al.(2025)Rana, Lee, Pershouse, and
  Suenderhauf]{rana2025imlepolicy}
Krishan Rana, Robert Lee, David Pershouse, and Niko Suenderhauf.
\newblock {IMLE} policy: Fast and sample efficient visuomotor policy learning
  via implicit maximum likelihood estimation.
\newblock \emph{arXiv preprint arXiv:2502.12371}, 2025.

\bibitem[Ross et~al.(2011)Ross, Gordon, and Bagnell]{ross2011dagger}
St{\'e}phane Ross, Geoffrey~J. Gordon, and Drew Bagnell.
\newblock A reduction of imitation learning and structured prediction to
  no-regret online learning.
\newblock In \emph{International Conference on Artificial Intelligence and
  Statistics (AISTATS)}, 2011.

\bibitem[Rubinstein(1999)]{rubinstein1999cem}
Reuven~Y. Rubinstein.
\newblock The cross-entropy method for combinatorial and continuous
  optimization.
\newblock \emph{Methodology and Computing in Applied Probability}, 1\penalty0
  (2):\penalty0 127--190, 1999.

\bibitem[Rupprecht et~al.(2017)Rupprecht, Laina, DiPietro, Baust, Tombari,
  Navab, and Hager]{rupprecht2017mhp}
Christian Rupprecht, Iro Laina, Robert DiPietro, Maximilian Baust, Federico
  Tombari, Nassir Navab, and Gregory~D. Hager.
\newblock Learning in an uncertain world: Representing ambiguity through
  multiple hypotheses.
\newblock In \emph{IEEE International Conference on Computer Vision (ICCV)},
  2017.

\bibitem[Scheuerer \& Hamill(2015)Scheuerer and Hamill]{scheuerer2015variogram}
Michael Scheuerer and Thomas~M. Hamill.
\newblock Variogram-based proper scoring rules for probabilistic forecasts of
  multivariate quantities.
\newblock \emph{Monthly Weather Review}, 143\penalty0 (4):\penalty0 1321--1334,
  2015.

\bibitem[Seo et~al.(2020)Seo, Lee, Clavera, Kurutach, Shin, and
  Abbeel]{seo2020tmcl}
Younggyo Seo, Kimin Lee, Ignasi Clavera, Thanard Kurutach, Jinwoo Shin, and
  Pieter Abbeel.
\newblock Trajectory-wise multiple choice learning for dynamics generalization
  in reinforcement learning.
\newblock In \emph{Advances in Neural Information Processing Systems
  (NeurIPS)}, 2020.

\bibitem[Shannon(1959)]{shannon1959coding}
Claude~E. Shannon.
\newblock Coding theorems for a discrete source with a fidelity criterion.
\newblock In \emph{IRE National Convention Record}, volume~4, pp.\  142--163,
  1959.

\bibitem[Sim{\'e}oni et~al.(2025)Sim{\'e}oni, Vo, Seitzer, Baldassarre, Oquab,
  Jose, Khalidov, Szafraniec, Yi, Ramamonjisoa, Massa, Haziza, Wehrstedt, Wang,
  Darcet, Moutakanni, Sentana, Roberts, Vedaldi, Tolan, Brandt, Couprie,
  Mairal, J{\'e}gou, Labatut, and Bojanowski]{simeoni2025dinov3}
Oriane Sim{\'e}oni, Huy~V. Vo, Maximilian Seitzer, Federico Baldassarre, Maxime
  Oquab, Cijo Jose, Vasil Khalidov, Marc Szafraniec, Seungeun Yi, Micha{\"e}l
  Ramamonjisoa, Francisco Massa, Daniel Haziza, Luca Wehrstedt, Jianyuan Wang,
  Timoth{\'e}e Darcet, Th{\'e}o Moutakanni, Leonel Sentana, Claire Roberts,
  Andrea Vedaldi, Jamie Tolan, John Brandt, Camille Couprie, Julien Mairal,
  Herv{\'e} J{\'e}gou, Patrick Labatut, and Piotr Bojanowski.
\newblock {DINOv3}.
\newblock \emph{arXiv preprint arXiv:2508.10104}, 2025.

\bibitem[Somepalli et~al.(2023)Somepalli, Singla, Goldblum, Geiping, and
  Goldstein]{somepalli2023diffusion}
Gowthami Somepalli, Vasu Singla, Micah Goldblum, Jonas Geiping, and Tom
  Goldstein.
\newblock Diffusion art or digital forgery? investigating data replication in
  diffusion models.
\newblock In \emph{IEEE/CVF Conference on Computer Vision and Pattern
  Recognition (CVPR)}, 2023.

\bibitem[Song et~al.(2026)Song, Xing, Tang, Huang, Huang, Yan, Lv, Yang, Niu,
  He, Wang, and Yao]{song2026branchjepa}
Zhi Song, Ximing Xing, Zhenchao Tang, Hanbo Huang, Jiehui Huang, Weilong Yan,
  Tianxu Lv, Minghao Yang, Zhongzheng Niu, Bing He, Lusheng Wang, and Jianhua
  Yao.
\newblock {Branch-JEPA}: Finite-support predictive distributions for {JEPA}
  world models.
\newblock \emph{arXiv preprint arXiv:2607.05238}, 2026.

\bibitem[Terver et~al.(2025)Terver, Yang, Ponce, Bardes, and
  LeCun]{terver2025drives}
Basile Terver, Tsung-Yen Yang, Jean Ponce, Adrien Bardes, and Yann LeCun.
\newblock What drives success in physical planning with joint-embedding
  predictive world models?
\newblock \emph{arXiv preprint arXiv:2512.24497}, 2025.

\bibitem[Thiede \& Brahma(2019)Thiede and Brahma]{thiede2019variety}
Luca~Anthony Thiede and Pratik~Prabhanjan Brahma.
\newblock Analyzing the variety loss in the context of probabilistic trajectory
  prediction.
\newblock In \emph{IEEE/CVF International Conference on Computer Vision
  (ICCV)}, 2019.

\bibitem[Vashist \& Li(2026)Vashist and Li]{vashist2026roms}
Chirag Vashist and Ke~Li.
\newblock {ROMS-IMLE}: A minimalist approach to competitive single-step
  generative modelling.
\newblock \emph{arXiv preprint arXiv:2607.19332}, 2026.

\bibitem[Vashist et~al.(2024)Vashist, Peng, and Li]{vashist2024rsimle}
Chirag Vashist, Shichong Peng, and Ke~Li.
\newblock Rejection sampling {IMLE}: Designing priors for better few-shot image
  synthesis.
\newblock \emph{arXiv preprint arXiv:2409.17439}, 2024.

\bibitem[Venkatraman et~al.(2015)Venkatraman, Hebert, and
  Bagnell]{venkatraman2015dad}
Arun Venkatraman, Martial Hebert, and J.~Andrew Bagnell.
\newblock Improving multi-step prediction of learned time series models.
\newblock In \emph{AAAI Conference on Artificial Intelligence}, 2015.

\bibitem[Vondrick et~al.(2016)Vondrick, Pirsiavash, and
  Torralba]{vondrick2016anticipating}
Carl Vondrick, Hamed Pirsiavash, and Antonio Torralba.
\newblock Anticipating visual representations from unlabeled video.
\newblock In \emph{IEEE Conference on Computer Vision and Pattern Recognition
  (CVPR)}, 2016.

\bibitem[Walker et~al.(2025)Walker, V{\'e}lez, Polania~Cabrera, Zhou, Kabra,
  Doersch, Ovsjanikov, Carreira, and Ginosar]{walker2025generalist}
Jacob~C. Walker, Pedro V{\'e}lez, Luisa Polania~Cabrera, Guangyao Zhou, Rishabh
  Kabra, Carl Doersch, Maks Ovsjanikov, Jo{\~a}o Carreira, and Shiry Ginosar.
\newblock Generalist forecasting with frozen video models via latent diffusion.
\newblock \emph{arXiv preprint arXiv:2507.13942}, 2025.

\bibitem[Wu et~al.(2026)Wu, Luo, Hariharan, Chao, and Campbell]{wu2026wtagmm}
Qiyuan Wu, Katie~Z. Luo, Bharath Hariharan, Wei-Lun Chao, and Mark Campbell.
\newblock Rethinking training \& inference for forecasting: Linking
  winner-take-all back to {GMMs}.
\newblock \emph{arXiv preprint arXiv:2606.26424}, 2026.

\bibitem[Yang et~al.(2019)Yang, Hong, Jang, Zhao, and Lee]{yang2019dsgan}
Dingdong Yang, Seunghoon Hong, Yunseok Jang, Tianchen Zhao, and Honglak Lee.
\newblock Diversity-sensitive conditional generative adversarial networks.
\newblock In \emph{International Conference on Learning Representations
  (ICLR)}, 2019.

\bibitem[Zhang et~al.(2025)Zhang, Karkus, Igl, Ding, Chen, Ivanovic, and
  Pavone]{zhang2025catk}
Zhejun Zhang, Peter Karkus, Maximilian Igl, Wenhao Ding, Yuxiao Chen, Boris
  Ivanovic, and Marco Pavone.
\newblock Closed-loop supervised fine-tuning of tokenized traffic models.
\newblock In \emph{IEEE/CVF Conference on Computer Vision and Pattern
  Recognition (CVPR)}, 2025.

\bibitem[Zheng et~al.(2025)Zheng, Ma, Tong, and Xie]{zheng2025rae}
Boyang Zheng, Nanye Ma, Shengbang Tong, and Saining Xie.
\newblock Diffusion transformers with representation autoencoders.
\newblock \emph{arXiv preprint arXiv:2510.11690}, 2025.

\bibitem[Zhou et~al.(2025)Zhou, Pan, LeCun, and Pinto]{zhou2025dinowm}
Gaoyue Zhou, Hengkai Pan, Yann LeCun, and Lerrel Pinto.
\newblock {DINO-WM}: World models on pre-trained visual features enable
  zero-shot planning.
\newblock In \emph{International Conference on Machine Learning (ICML)}, 2025.

\end{thebibliography}
\end{document}